\documentclass[12pt,twoside]{article}
\usepackage[dvips]{graphicx}
\usepackage{latexsym,amssymb,textcomp,wasysym,rotating}
\newtheorem{nummer}{\hspace*{-0.33em}}[section]

\newenvironment{Example} {\begin{nummer} {\bf Example.}
\begin{rm}} {\end{rm} \end{nummer}}
\newenvironment{keywords}{\centerline{\bf\small
Keywords}\begin{quote}\small}{\par\end{quote}\vskip 1ex}

\def\AA{\mathcal A}  
  
  \def\OO{\mathcal O}
\def\RR{\mathcal R}  \def\UU{\mathcal U}
  
 \def\PP{\mathcal P}

\def\NNN{\mathbb N}  \def\QQQ{\mathbb Q}

\def\tC{$\cdot$}    
\def\tB{$\bullet$}
\def\tA{$\newmoon$}  

\def\E{\mathbf{E}}              
\def\eps{\varepsilon}
\begin{document}

\title{\bf\Large\hrule height5pt \vskip 4mm
{\Huge Universal Intelligence:\\} A Definition of Machine Intelligence
\vskip 4mm \hrule height2pt}

\pagestyle{headings}

\author{\\
{\bf Shane Legg}\\[3mm]
\normalsize IDSIA, Galleria 2, Manno-Lugano CH-6928, Switzerland \\
\normalsize \texttt{shane@vetta.org \ \ \ www.vetta.org/shane}
\and \\
{\bf Marcus Hutter}\\[3mm]
\normalsize RSISE$\,$@$\,$ANU and SML$\,$@$\,$NICTA,
\normalsize Canberra, ACT, 0200, Australia \\
\normalsize \texttt{marcus@hutter1.net \ \ \ www.hutter1.net} \\[3ex]
}
\date{December 2007}
\maketitle

\begin{abstract}
A fundamental problem in artificial intelligence is that nobody really
knows what intelligence is.  The problem is especially acute when we
need to consider artificial systems which are significantly different
to humans.  In this paper we approach this problem in the following
way: We take a number of well known informal definitions of human
intelligence that have been given by experts, and extract their
essential features.  These are then mathematically formalised to
produce a general measure of intelligence for arbitrary machines.  We
believe that this equation formally captures the concept of machine
intelligence in the broadest reasonable sense.  We then show how this
formal definition is related to the theory of universal optimal
learning agents.  Finally, we survey the many other tests and
definitions of intelligence that have been proposed for machines.
\end{abstract}


\newpage \thispagestyle{empty}
\tableofcontents

\newpage
\section{Introduction}

\begin{quote}\it
``Innumerable tests are available for measuring intelligence, yet no
one is quite certain of what intelligence is, or even just what it is
that the available tests are measuring.''  \hfill --- {\sl R. L. Gregory}
\cite{Gregory:98}
\end{quote}

What is intelligence?  It is a concept that we use in our daily lives
that seems to have a fairly concrete, though perhaps naive, meaning.
We say that our friend who got an A in his calculus test is very
intelligent, or perhaps our cat who has learnt to go into hiding at
the first mention of the word ``vet''.  Although this intuitive notion
of intelligence presents us with no difficulties, if we attempt to dig
deeper and define it in precise terms we find the concept to be very
difficult to nail down.  Perhaps the ability to learn quickly is
central to intelligence?  Or perhaps the total sum of one's knowledge
is more important?  Perhaps communication and the ability to use
language play a central role?  What about ``thinking'' or the ability
to perform abstract reasoning?  How about the ability to be creative
and solve problems?  Intelligence involves a perplexing mixture of
concepts, many of which are equally difficult to define.

Psychologists have been grappling with these issues ever since humans
first became fascinated with the nature of the mind.  Debates have
raged back and forth concerning the correct definition of intelligence
and how best to measure the intelligence of individuals.  These
debates have in many instances been very heated as what is at stake is
not merely a scientific definition, but a fundamental issue of how we
measure and value humans: Is one employee smarter than another?  Are
men on average more intelligent than women?  Are white people smarter
than black people?  As a result intelligence tests, and their
creators, have on occasion been the subject of intense public
scrutiny.  Simply determining whether a test, perhaps quite
unintentionally, is partly a reflection of the race, gender, culture
or social class of its creator is a subtle, complex and often
politically charged issue \cite{Gould:81,Herrnstein:96}.  Not
surprisingly, many have concluded that it is wise to stay well clear
of this topic.

In reality the situation is not as bad as it is sometimes made out to
be.  Although the details of the definition are debated, in broad
terms a fair degree of consensus about the scientific definition of
intelligence and how to measure it has been achieved
\cite{Gottfredson:97msoi,Sternberg:86}.  Indeed it is widely
recognised that when standard intelligence tests are correctly applied
and interpreted, they all measure approximately the same thing
\cite{Gottfredson:97msoi}.  Furthermore, what they measure is both
stable over time in individuals and has significant predictive power,
in particular for future academic performance and other mentally
demanding pursuits.  The issues that continue to draw debate are the
questions such as whether the tests test only a part or a particular
type of intelligence, or whether they are somehow biased towards a
particular group or set of mental skills.  Great effort has gone into
dealing with these issues, but they are difficult problems with no
easy solutions.

Somewhat disconnected from this exists a parallel debate over the
nature of intelligence in the context of machines.  While the debate
is less politically charged, in some ways the central issues are even
more difficult.  Machines can have physical forms, sensors, actuators,
means of communication, information processing abilities and
environments that are totally unlike those that we experience.  This
makes the concept of ``machine intelligence'' particularly difficult
to get a handle on.  In some cases, a machine may display properties
that we equate with human intelligence, in such cases it might be
reasonable to describe the machine as also being intelligent.  In
other situations this view is far too limited and anthropocentric.
Ideally we would like to be able to measure the intelligence of a wide
range of systems; humans, dogs, flies, robots or even disembodied
systems such as chat-bots, expert systems, classification systems and
prediction algorithms \cite{Johnson:92,Albus:91}.

One response to this problem might be to develop specific kinds of
tests for specific kinds of entities; just as intelligence tests for
children differ to intelligence tests for adults.  While this works
well when testing humans of different ages, it comes undone when we
need to measure the intelligence of entities which are profoundly
different to each other in terms of their cognitive capacities, speed,
senses, environments in which they operate, and so on.  To measure the
intelligence of such diverse systems in a meaningful way we must step
back from the specifics of particular systems and establish the
underlying fundamentals of what it is that we are really trying to
measure.

The difficulty of developing an abstract and highly general notion of
intelligence is readily apparent.  Consider, for example, the memory
and numerical computation tasks that appear in some intelligence tests
and which were once regarded as defining hallmarks of human
intelligence.  We now know that these tasks are absolutely trivial for
a machine and thus do not appear to test the machine's intelligence in
any meaningful sense.  Indeed even the mentally demanding task of
playing chess can be largely reduced to brute force search
\cite{Hsu:95}.  What else may in time be possible with relatively
simple algorithms running on powerful machines is hard to say.  What
we can be sure of is that as technology advances, our concept of
intelligence will continue to evolve with~it.

How then are we to develop a concept of intelligence that is
applicable to all kinds of systems?  Any proposed definition must
encompass the essence of human intelligence, as well as other
possibilities, in a consistent way.  It should not be limited to any
particular set of senses, environments or goals, nor should it be
limited to any specific kind of hardware, such as silicon or
biological neurons.  It should be based on principles which are
fundamental and thus unlikely to alter over time.  Furthermore, the
definition of intelligence should ideally be formally expressed,
objective, and practically realisable as an effective intelligence
test.

\vspace{0.5em}

In this paper we approach the problem of defining machine intelligence
as follows:\vspace{0.3em}\\ \emph{Section~\ref{sec:ni}} overviews well
known theories, definitions and tests of intelligence that have been
developed by psychologists.  Our objective in this section is to gain
an understanding of the essence of intelligence in the broadest
possible terms.  In particular we are interested in commonly expressed
ideas that could be applied to arbitrary systems and contexts, not
just humans.\vspace{0.3em}\\ \emph{Section~\ref{sec:fmi}} takes these
key ideas and formalises them.  This leads to \emph{universal
  intelligence}, our proposed formal definition of machine
intelligence.  We then examine some of the properties of universal
intelligence, such as its ability to sensibly order simple learning
algorithms and connections to the theory of universal optimal learning
agents.\vspace{0.3em}\\ \emph{Section~\ref{sec:ai}} overviews other
definitions and tests of machine intelligence that have been proposed.
Although surveys of the Turing test and its many variants exist, for
example \cite{Saygin:00}, as far as we know this section is the first
general survey of definitions and tests of machine intelligence.
Given how fundamental this is to the field of artificial intelligence,
the absence of such a survey is quite remarkable.  For any field to
mature as a science, questions of definition and measurement must be
meticulously investigated.  We conclude our survey with a summary
comparison of the various proposed tests and definitions of machine
intelligence.
\vspace{0.3em}\\ \emph{Section~\ref{sec:conc}} ends the paper with
discussion, responses to criticisms, conclusions and directions for
future research.

\vspace{0.5em}

The genesis of this work lies in Hutter's universal optimal learning
agent, AIXI, described in 2, 12, 60 and 300 pages in
\cite{Hutter:01decision,Hutter:01aixi, Hutter:04uaibook,
  Hutter:07aixigentle}, respectively.  In this work, an order relation
for intelligent agents is presented, with respect to which the
provably optimal AIXI agent is maximal.  The universal intelligence
measure presented here is a derivative of this order relation.  A
short description of the universal intelligence measure appeared in
\cite{Legg:05iors}, from which two articles followed in the popular
scientific press \cite{Graham-Rowe:05,Fievet:05}.  An 8 page paper on
universal intelligence appeared in \cite{Legg:06ior}, followed by an
updated poster presentation \cite{Legg:06iors}.  In the current paper
we explore universal intelligence in much greater detail, in
particular the way in which it relates to mainstream views on human
intelligence and other proposed definitions of machine intelligence.

\section{Natural Intelligence}\label{sec:ni}

Human intelligence is an enormously rich topic with a complex
intellectual, social and political history.  For an overview the
interested reader might want to consult ``Handbook of Intelligence''
\cite{Sternberg:00} edited by R. J. Sternberg.  Our objective in this
section is simply to sketch a range of tests, theories and definitions
of human and animal intelligence.  We are particularly interested in
common themes and general perspectives on intelligence that could be
applicable to many kinds of systems, as these will form the foundation
of our definition of machine intelligence in the next section.

\subsection{Human intelligence tests}
\label{subsec:sti}

Contrary to popular public opinion, most psychologists believe that
the usual tests of intelligence, such as IQ tests, reliably measure
something important in humans \cite{Neisser:96,Gottfredson:97}.  In
fact, standard intelligence tests are among the most statistically
stable and reliable of psychological tests.  Furthermore, it is well
known that these scores are a good predictor of various things, such
as academic performance.  The question then is not whether these tests
are useful or measure something meaningful, but rather whether what
they measure is indeed ``intelligence''.  Some experts believe that
they do, while others think that they only succeed in measuring
certain aspects of, or types of, intelligence.

The first modern style intelligence test was developed by the French
psychologist Alfred Binet in 1905.  Binet believed that intelligence
was best studied by looking at relatively complex mental tasks, unlike
earlier tests developed by Francis Galton which focused on reaction
times, auditory discrimination ability, physical coordination and so
on.  Binet's test consisted of 30 short tasks related to everyday
problems such as; naming parts of the body, comparing lengths and
weights, counting coins, remembering digits and definitions of words.
For each task category there were a number of problems of increasing
difficulty.  The child's results were obtained by normalising their
raw score against peers of the same age.  Initially his test was
designed to measure the mental performance of children with learning
problems \cite{Binet:1905}.  Later versions were also developed for
normal children \cite{Binet:1911}.  It was found that Binet's test
results were a good predictor of children's academic performance.

Lewis Terman of Stanford University developed an English version of
Binet's test.  As the age norms for French children did not correspond
well with American children, he revised Binet's test in various ways,
in particular he increased the upper age limit.  This resulted in the
now famous Stanford-Binet test \cite{Terman:50}.  This test formed the
basis of a number of other intelligence tests, such as the Army Alpha
and Army Beta tests which were used to classify recruits.  Since its
development, the Stanford-Binet has been periodically revised, with
updated versions being widely used today.

David Wechsler believed that the original Binet tests were too focused
on verbal skills and thus disadvantaged certain otherwise intelligent
individuals, for example the deaf or people who did not speak the test
language as a first language.  To address this problem, he proposed
that tests should contain a combination of both verbal and nonverbal
problems.  He also believed that in addition to an overall IQ score, a
profile should be produced showing the performance of the individual
in the various areas tested.  Borrowing significantly from the
Stanford-Binet, the US army Alpha test, and others, he developed a
range of tests targeting specific age groups from preschoolers up to
adults~\cite{Wechsler:58}.  Due in part to problems with revisions of
the Stanford-Binet test in the 1960's and 1970's, Wechsler's tests
became the standard.  They continue to be well respected and widely
used.

Owing to a common lineage, modern versions of the Wechsler and the
Stanford-Binet have a similar basic structure~\cite{Kaufman:00}.  Both
test the individual in a number of verbal and non-verbal ways.  In the
case of a Stanford-Binet the test is broken up into 5 key areas: Fluid
reasoning, knowledge, quantitative reasoning, visual-spatial
processing, and working memory.  In the case of the Wechsler Adult
Intelligence Scale (WAIS-III), the verbal tests include areas such as
such as knowledge, basic arithmetic, comprehension, vocabulary, and
short term memory.  Non-verbal tests include picture completion,
spatial perception, problem solving, symbol search and object
assembly.

As part of an effort to make intelligence tests more culture neutral
John Raven developed the progressive matrices test \cite{Raven:00}.
In this test each problem consists of a short sequence of basic
shapes.  For example, a circle in a box, then a circle with a cross in
the middle followed by a circle with a triangle inside.  The test
subject then has to select from a second list the image that best
continues the pattern.  Simple problems have simple patterns, while
difficult problems have more subtle and complex patterns.  In each
case however, the simplest pattern that can explain the observed
sequence is the one that correctly predicts its continuation.  Thus,
not only is the ability to recognise patterns tested, but also the
ability to evaluate the complexity of different explanations and then
correctly apply the philosophical principle of Occam's razor.  We will
return to Occam's razor and its importance in intelligence testing in
Subsection~\ref{subsec:fmi} when considering machine intelligence.

Today several different versions of the Raven test exist designed for
different age groups and ability levels.  As the tests depend strongly
on the ability to identify abstract patterns, rather than knowledge,
they are considered to be some of the most ``g-loaded'' intelligence
tests available (see Subsection~\ref{subsec:toi}).  The Raven tests
remain in common use today, particularly when it is thought that
culture or language bias could be an issue.

The intelligence quotient, or IQ, was originally introduced by Stern
\cite{Stern:1912}.  It was computed by taking the age of a child as
estimated by their performance in the intelligence test, and then
dividing this by their true biological age and multiplying by 100.
Thus a 10 year old child whose mental performance was equal to that of
a normal 12 year old, had an IQ of 120.  As the concept of mental age
has now been discredited, and was never applicable to adults anyway,
modern IQ scores are simply normalised to a Gaussian distribution with
a mean of 100.  The standard deviation used varies: in the United
States 15 is commonly used, while in Europe 25 is common.  For
children the normalising Gaussian is based on peers of the same age.

Whatever normalising distribution is used, by definition an
individual's IQ is always an indication of their cognitive performance
relative to some larger group.  Clearly this would be problematic in
the context of machines where the performance of some machines could
be many orders of magnitude greater than others.  Furthermore, the
distribution of machine performance would be continually changing due
to advancing technology.  Thus, for our purposes, an absolute measure
will be more meaningful than a traditional IQ type of measure.

For an overview of the history of intelligence testing and the
structure of modern tests, see~\cite{Kaufman:00}.

\subsection{Animal intelligence tests}
\label{subsec:animal}

Testing the intelligence of animals is of particular interest to us as
it moves beyond strictly human focused concepts of intelligence and
testing methods.  Difficult problems in human intelligence testing,
such as bias due to language differences or physical handicap, become
even more difficult if we try to compare animals with different
perceptual and cognitive capacities.  Even within a single species
measurement is difficult as it is not always obvious how to conduct
the tests, or even what should be tested for.  Furthermore, as humans
devise the tests, there is a persistent danger that the tests may be
biased in terms of our sensory, motor, and motivational systems
\cite{Macphail:85}.  For example, it is known that rats can learn some
types of relationships much more easily through smell rather than
other senses \cite{Slotnick:74}.  Furthermore, while an IQ test for
children might in some sense be validated by its ability to predict
future academic or other success, it is not always clear how to
validate an intelligence test for animals.  If survival or the total
number of offspring was a measure of success, then bacteria might the
be most intelligent life on earth!

As is often the case when we try to generalise concepts, abstraction
is necessary.  When attempting to measure the intelligence of lower
animals it is necessary to focus on simple things like short and long
term memory, the forming of associations, the ability to generalise
simple patterns and make predictions, simple counting and basic
communication.  It is only with relatively intelligent social animals,
such as birds and apes, that more sophisticated properties such as
deception, imitation and the ability to recognise oneself become
important.  For simpler animals, the focus is more on the animal's
essential information processing capacity.  For example, the work on
understanding the capacity of ants to remember patterns when retracing
a path back to a source of food without the aid of
pheromones~\cite{Reznikova:86}.

One interesting difficulty when testing animal intelligence is that we
are unable to directly explain to the animal what its goal is.
Instead, we have to guide the animal towards a problem by carefully
rewarding selected behaviours with something like food.  In general,
when testing machine intelligence we face a similar problem in that we
cannot assume that a machine will have a sufficient level of language
comprehension to be able to understand commands.  Thus a simple
solution is to use basic ``rewards'' to guide behaviour, as we do with
animals.  Although this approach is extremely general, one difficulty
is that solving the task, and simply learning what the task is, become
confounded and thus the results need to be interpreted
carefully~\cite{Zentall:97}.  Due to our need for generality, we will
use this reward based approach for our formal measure of machine
intelligence.  Specifically, we will adopt the reinforcement learning
framework from artificial intelligence (see
Subsection~\ref{subsec:aef}).

For good overviews of animal intelligence research see
\cite{Zentall:00} or \cite{Herman:94}.

\subsection{Desirable properties of an intelligence test}
\label{subsec:desirable}

There are many properties that a good test of human intelligence
should have.  One important property is that the test should be
repeatable, in the sense that it consistently returns about the same
score for a given individual.  For example, the test subject should
not be able to significantly improve their performance if tested again
a short time later.  Statistical variability can also be a problem in
short tests.  Longer tests help in this regard, however they are
naturally more costly to administer.

Another important reliability factor is the bias that might be
introduced by the individual administering the test.  Purely written
tests avoid this problem as there is minimal interaction between the
tested individual and the tester.  However this lack of interaction
also has disadvantages as it may mean that other sources of bias, such
as cultural differences, language problems or even something as simple
as poor eyesight, might not be properly identified.  Thus, even in a
written test the individual being tested should first be examined by
an expert in order to ensure that the test is appropriate.

Cultural bias in particular is a difficult problem, and tests should
be designed to minimise this problem where possible, or at least
detect potential bias problems when they occur.  One way to do this is
to test each ability in multiple ways, for example both verbally and
visually.  While language is an obvious potential source of cultural
bias, more subtle forms of bias are difficult to detect and remedy.
For example, different cultures emphasise different cognitive
abilities, and thus it is difficult, perhaps impossible, to compare
intelligence scores in a way that is truly objective.  In part this is
a question of what intelligence is.  Indeed the problem of how to
weight performance in different areas is fundamental and we will need
to face it again in the context of our formal definition of machine
intelligence.

When testing large numbers of individuals, for example when testing
army recruits, the cost of administering the test becomes important.
In these cases less accurate but more economical test procedures may
be used, such as purely written tests without any direct interaction
between the individuals being tested and a psychologist.

An intelligence test should be valid in the sense that it appears to
be testing what it claims it is testing for.  One way to check this is
to show that the test produces results consistent with other
manifestations of intelligence.  A test should also have predictive
power, for example the ability to predict future academic performance.
This ensures that what is being measured is somehow meaningful, beyond
just the ability to answer the questions in the test.

Standard intelligence tests such as a modern Stanford-Binet are
thoroughly tested for years on the above criteria, and many others,
before they are ready for wide spread use.  Many of these desirable
properties, such as reliability, tester bias, cost and validity, are
also relevant to tests of machine intelligence.  To some extent they
are also relevant to formal definitions of intelligence.  We will
return to these desirable properties when analysing our definition of
machine intelligence in Subsection~\ref{subsec:anal}, and later when
comparing tests of machine intelligence in
Subsection~\ref{subsec:despropmi}.

\subsection{Static vs.\ dynamic tests}
\label{subsec:dynamic}

Stanford-Binet, Wechsler, Raven progressive matrices, and indeed most
standard intelligence tests, are all examples of ``static tests''.  By
this we mean that they test an individual's knowledge and ability to
solve one-off problems.  They do not directly measure the ability to
learn and adapt over time.  If an individual was good at learning and
adapting then we might expect this to be reflected in their total
knowledge and thus picked up in a static test.  However, it could be
that an individual has a great capacity to learn, but that this is not
reflected in their knowledge due to limited education.  In which case,
if we consider the capacity to learn and adapt rather than the sum of
knowledge to be a defining characteristic of intelligence, then to
class an individual as unintelligent due to limited access to
education would be a mistake.

What is needed is a more direct test of an individual's ability to
learn and adapt: A so called ``dynamic test''\cite{Sternberg:02} (for
related work see also~\cite{Johnson-Laird:77}).  In a dynamic test the
test subject interacts over a period of time with the tester, who now
becomes a kind of teacher.  The tester's task is to present the
individual with a series of problems.  After each attempt at solving a
problem, the tester provides feedback to the individual who then has
to adapt their behaviour accordingly in order to solve the next
problem.

Although dynamic tests could in theory be very powerful, they are not
yet well established due to a number of difficulties.  One of the
drawbacks is that they require a much greater degree of interaction
between the test subject and the tester.  This makes dynamic testing
more costly to perform and increases the danger of tester bias.

Dynamic testing is of particular interest to us because in a formal
test for machines it appears that we can overcome these problems by
automating the role of the tester.

\subsection{Theories of human intelligence}
\label{subsec:toi}

Complementary to the experimental study of human intelligence,
theories have been developed that attempt to better characterise the
fundamental nature of intelligence.  It is useful for us to briefly
sketch this work as some of these issues have parallels within the
context of machine intelligence.

One central question is whether intelligence should be viewed as one
ability, or many.  On one side of the debate are the theories that
view intelligence as consisting of many different components and that
identifying these components is important to understanding
intelligence.  Different theories propose different ways to do this.
One of the first was Thurstone's ``multiple-factors'' theory which
considers seven ``primary mental abilities'': verbal comprehension,
word fluency, number facility, spatial visualisation, associative
memory, perceptual speed and reasoning \cite{Thurstone:38}.  Another
approach is Sternberg's ``Triarchic Mind'' which breaks intelligence
down into analytical intelligence, creative intelligence, and
practical intelligence \cite{Sternberg:85}, however this model is now
considered outdated, even by Sternberg himself.

Taking the number of components to an extreme is Guilford's
``Structure of Intellect'' theory.  Under this theory there are three
fundamental dimensions: Contents, operations, and products.  Together
these give rise to 120 different categories \cite{Guilford:67} (in
later work this increased to 150 categories).  This theory has been
criticised due to the fact that measuring such precise combinations of
cognitive capacities in individuals seems to be infeasible and thus it
is difficult to experimentally study such a fine grained model of
intelligence.

A recently popular approach is Gardner's ``multiple intelligences''
where he argues that the components of human intelligence are
sufficiently separate that they are actually different
``intelligences''\cite{Gardner:93}.  Based on the structure of the
human brain he identifies these intelligences to be linguistic,
musical, logical-mathematical, spatial, bodily kinaesthetic,
intra-personal and inter-personal intelligence.  Although Gardner's
theory of multiple intelligences has certainly captured the
imagination of the public, it remains to be seen to what degree it
will have a lasting impact in professional circles.

At the other end of the spectrum is the work of Spearman and those
that have followed in his footsteps.  Here intelligence is seen as a
very general mental ability that underlies and contributes to all
other mental abilities.  As evidence they point to the fact that an
individual's performance levels in reasoning, association, linguistic,
spatial thinking, pattern identification etc.\ are positively
correlated.  Spearman called this positive statistical correlation
between different mental abilities the ``$g$-factor'', where $g$
stands for ``general intelligence''\cite{Spearman:27}.  Because
standard IQ tests measure a range of key cognitive abilities, from a
collection of scores on different cognitive tasks we can estimate an
individual's $g$-factor.  Some who consider the generality of
intelligence to be of primary importance take the $g$-factor to be the
very definition of intelligence~\cite{Gottfredson:02}.

A well known refinement to the $g$-factor theory due to Cattell is to
distinguish between, ``fluid intelligence'', which is a very general
and flexible innate ability to deal with problems and complexity, and
``crystallized intelligence'', which measures the knowledge and
abilities that an individual has acquired over time \cite{Cattell:87}.
For example, while an adolescent may have a similar level of fluid
intelligence to that of an adult, their level of crystallized
intelligence is typically lower due to less life experience
\cite{Horn:70}.  Although it is difficult to determine to what extent
these two influence each other, the distinction is an important one
because it captures two distinct notions of what the word
``intelligence'' means.

As the $g$-factor is simply the statistical correlation between
difference kinds of mental abilities, it is not fundamentally
inconsistent with the view that intelligence can have multiple aspects
or dimensions.  Thus a synthesis of the two perspectives is possible
by viewing intelligence as a hierarchy with the $g$-factor at its apex
and increasing levels of specialisation for the different aspects of
intelligence forming branches \cite{Carroll:93}.  For example, an
individual might have a high $g$-factor, which contributes to all of
their cognitive abilities, but also have an especially well developed
musical sense.  This hierarchical view of intelligence is now quite
popular \cite{Neisser:96}.

\subsection{Ten definitions of human intelligence}
\label{subsec:idi}

\begin{quote}\it
``Viewed narrowly, there seem to be almost as many definitions of
intelligence as there were experts asked to define it.''
  \hfill --- {\sl R. J. Sternberg quoted in~\cite{Gregory:98}}
\end{quote}

In this subsection and the next we will overview a range of
definitions of intelligence that have been given by psychologists.
Many of these definitions are well known.  Although the definitions
differ, there are reoccurring features; in some cases these are
explicitly stated, while in others they are more implicit.  We start
by considering ten definitions that take, in our view, a similar
perspective:

\begin{itemize}

\item[] ``It seems to us that in intelligence there is a fundamental
  faculty, the alteration or the lack of which, is of the utmost
  importance for practical life. This faculty is judgement, otherwise
  called good sense, practical sense, initiative, the faculty of
  adapting oneself to circumstances.''
  \mbox{A. Binet~\cite{Binet:1905}}

\item[] ``The capacity to learn or to profit by experience.''
  W. F. Dearborn quoted in~\cite{Sternberg:00}

\item[] ``Ability to adapt oneself adequately to relatively new
  situations in life.''  R.~Pinter quoted in~\cite{Sternberg:00}

\item[] ``A person possesses intelligence insofar as he has learned,
  or can learn, to adjust himself to his environment.''  S. S. Colvin
  quoted in~\cite{Sternberg:00}

\item[] ``We shall use the term `intelligence' to mean the ability of
  an organism to solve new problems \ldots''
  W. V. Bingham~\cite{Bingham:37}

\item[] ``A global concept that involves an individual's ability to
  act purposefully, think rationally, and deal effectively with the
  environment.''  D. Wechsler~\cite{Wechsler:58}

\item[] ``Individuals differ from one another in their ability to
  understand complex ideas, to adapt effectively to the environment,
  to learn from experience, to engage in various forms of reasoning,
  to overcome obstacles by taking thought.''  American Psychological
  Association~\cite{Neisser:96}

\item[] ``\ldots I prefer to refer to it as `successful intelligence.'
  And the reason is that the emphasis is on the use of your
  intelligence to achieve success in your life.  So I define it as
  your skill in achieving whatever it is you want to attain in your
  life within your sociocultural context --- meaning that people have
  different goals for themselves, and for some it's to get very good
  grades in school and to do well on tests, and for others it might be
  to become a very good basketball player or actress or musician.''
  R. J. Sternberg~\cite{Sternberg:03}

\item[] ``Intelligence is part of the internal environment that shows
  through at the interface between person and external environment as
  a function of cognitive task demands.'' R. E. Snow quoted
  in~\cite{Slatter:01}

\item[] ``\ldots certain set of cognitive capacities that enable an
  individual to adapt and thrive in any given environment they find
  themselves in, and those cognitive capacities include things like
  memory and retrieval, and problem solving and so forth. There's a
  cluster of cognitive abilities that lead to successful adaptation to
  a wide range of environments.'' D. K. Simonton~\cite{Simonton:03}

\end{itemize}

Perhaps the most elementary common feature of these definitions is
that intelligence is seen as a property of an individual who is
interacting with an external environment, problem or situation.
Indeed, at least this much is common to practically all proposed
definitions of intelligence.

Another common feature is that an individual's intelligence is related
to their ability to succeed or ``profit''.  The notion of success or
profit implies the existence of some kind of objective or goal.  What
the goal is, is not specified, indeed individuals' goals may be
varied.  The important thing is that the individual is able to
carefully choose their actions in a way that leads to them
accomplishing their goals.  The greater this capacity to succeed with
respect to various goals, the greater the individual's intelligence.

The strong emphasis on learning, adaption and experience in these
definitions implies that the environment is not fully known to the
individual and may contain new situations that could not have been
anticipated in advance.  Thus intelligence is not the ability to deal
with a fully known environment, but rather the ability to deal with
some range of possibilities which cannot be wholly anticipated.  What
is important then is that the individual is able to quickly learn and
adapt so as to perform as well as possible over a wide range of
environments, situations, tasks and problems.  Collectively we will
refer to these as ``environments'', similar to some of the definitions
above.

Bringing these key features together gives us what we believe to the
essence of intelligence in its most general form:
\begin{quote}
\emph{Intelligence measures an agent's ability to achieve goals in a
  wide range of environments.}
\end{quote}

We take this to be our informal working definition of intelligence.
In the next section we will use this definition as the starting point
from which we will construct a formal definition of machine
intelligence.  However before we proceed further, the reader way wish
to revise the 10 definitions above to ensure that the definition we
have adopted is indeed reasonable.

\subsection{More definitions of human intelligence}
\label{subsec:mdi}

Of course many other definitions of intelligence have been proposed
over the years.  Usually they are not entirely incompatible with our
informal definition, but rather stress different aspects of
intelligence.  In this subsection we will survey some of these other
definitions and compare them to the position we have taken.  For an
even more extensive collection of definitions of intelligence, indeed
the largest collection that we are aware of, visit our online
collection~\cite{Legg:07idefs}.

The following is an especially interesting definition as it was given
as part of a group statement signed by 52 experts in the field.  As
such it obviously represents a fairly mainstream perspective:

\begin{itemize}
\item[] ``Intelligence is a very general mental capability that, among
  other things, involves the ability to reason, plan, solve problems,
  think abstractly, comprehend complex ideas, learn quickly and learn
  from experience.'' \cite{Gottfredson:97msoi}
\end{itemize}

Reasoning, planning, solving problems, abstract thinking, learning
from experience and so on, these are all mental abilities that allow
us to successfully achieve goals.  If we were missing any one of these
capacities, we would clearly be less able to successfully deal with
such a wide range of environments.  Thus, these capacities are
implicit in our definition also.  The difference is that our
definition does not attempt to specify what capabilities might be
needed, something which is clearly very difficult and would depend on
the particular tasks that the agent must deal with.  Our approach is
to consider intelligence to be the \emph{effect} of capacities such as
those listed above.  It is not the result of having any specific set
of capacities.  Indeed, intelligence could also be the effect of many
other capacities, some of which humans may not have.  In summary, our
definition is not in conflict with the above definition, rather it is
that our definition is more abstract and general.

\begin{itemize}
\item[] ``\ldots in its lowest terms intelligence is present where the
  individual animal, or human being, is aware, however dimly, of the
  relevance of his behaviour to an objective.  Many definitions of
  what is indefinable have been attempted by psychologists, of which
  the least unsatisfactory are 1. the capacity to meet novel
  situations, or to learn to do so, by new adaptive responses and
  2. the ability to perform tests or tasks, involving the grasping of
  relationships, the degree of intelligence being proportional to the
  complexity, or the abstractness, or both, of the relationship.''
  J. Drever~\cite{Drever:52}
\end{itemize}

This definition has many similarities to ours.  Firstly, it emphasises
the agent's ability to choose its actions so as to achieve an
objective, or in our terminology, a goal.  It then goes on to stress
the agent's ability to deal with situations which have not been
encountered before.  In our terminology, this is the ability to deal
with a wide range of environments.  Finally, this definition
highlights the agent's ability to perform tests or tasks, something
which is entirely consistent with our performance orientated
perspective of intelligence.

\begin{itemize}
\item[] ``Intelligence is not a single, unitary ability, but rather a
  composite of several functions. The term denotes that combination of
  abilities required for survival and advancement within a particular
  culture.'' A. Anastasi~\cite{Anastasi:92}
\end{itemize}

This definition does not specify exactly which capacities are
important, only that they should enable the individual to survive and
advance with the culture.  As such this is a more abstract ``success''
orientated definition of intelligence, like ours.  Naturally, culture
is a part of the agent's environment.

\begin{itemize}
\item[] ``The ability to carry on abstract thinking.''  L. M. Terman
  quoted in~\cite{Sternberg:00}
\end{itemize}

This is not really much of a definition as it simply shifts the
problem of defining intelligence to the problem of defining abstract
thinking.  The same is true of many other definitions that refer to
things such as imagination, creativity or consciousness.  The
following definition has a similar problem:

\begin{itemize}
\item[] ``The capacity for knowledge, and knowledge possessed.''
  V. A. C. Henmon \cite{Henmon:21}
\end{itemize}

What exactly constitutes ``knowledge'', as opposed to perhaps data or
information?  For example, does a library contain a lot of knowledge,
and if so, is it intelligent?  Or perhaps the internet?  Modern
concepts of the word knowledge stress the fact that the information
has to be in some sense properly contextualised so that it has
meaning.  Defining this more precisely appears to be difficult
however.  Because this definition of intelligence dates from 1921,
perhaps it reflects pre-information age thinking when computers with
vast storage capacities did not exist.

Nonetheless, our definition of intelligence is not entirely
inconsistent with the above definition in that an individual may be
required to know many things, or have a significant capacity for
knowledge, in order to perform well in some environments.  However our
definition is broader in that knowledge, or the capacity for
knowledge, is not by itself sufficient.  We require that the knowledge
can be used effectively for some purpose.  Indeed unless information
can be effectively utilised for a number of purposes, it seems
reasonable to consider it to be merely ``data'', rather than
``knowledge''.

\begin{itemize}
\item[] ``The capacity to acquire capacity.''  H. Woodrow quoted
  in~\cite{Sternberg:00}
\end{itemize}

The definition of Woodrow is typical of those which emphasise not the
current ability of the individual, but rather the individual's ability
to expand and develop new abilities.  This is a fundamental point of
divergence for many views on intelligence.  Consider the following
question: Is a young child as intelligent as an adult?  From one
perspective, children are very intelligent because they can learn and
adapt to new situations quickly.  On the other hand, the child is
unable to do many things due to a lack of knowledge and experience and
thus will make mistakes an adult would know to avoid.  These need not
just be physical acts, they could also be more subtle things like
errors in reasoning as their mind, while very malleable, has not yet
matured.  In which case, perhaps their intelligence is currently low,
but will increase with time and experience?

Fundamentally, this difference in perspective is a question of time
scale: Must an agent be able to tackle some task immediately, or
perhaps after a short period of time during which learning can take
place, or perhaps it only matters that they can eventually learn to
deal with the problem?  Being able to deal with a difficult problem
immediately is a matter of experience, rather than intelligence.
While being able to deal with it in the very long run might not
require much intelligence at all, for example, simply trying a vast
number of possible solutions might eventually produce the desired
results.  Intelligence then seems to be the ability to adapt and learn
as quickly as possible given the constraints imposed by the problem at
hand.  It is this insight that we will use to neatly deal with
temporal preference when defining machine intelligence (see
\emph{Measure of success} in Subsection~\ref{subsec:formframe}).

\begin{itemize}
\item[] ``Intelligence is a general factor that runs through all
types of performance.'' A.~Jensen
\end{itemize}

At first this might not look like a definition of intelligence, but it
makes an important point: Intelligence is not really the ability to do
anything in particular, rather it is a very general ability that
affects many kinds of performance.  Conversely, by measuring many
different kinds of performance we can estimate an individual's
intelligence.  This is consistent with our definition's emphasis on
the agent's generality.

\begin{itemize}
\item[] ``Intelligence is what is measured by intelligence tests.''
E. Boring~\cite{Boring:23}
\end{itemize}

Boring's famous definition of intelligence takes this idea a step
further.  If intelligence is not the ability to do anything in
particular, but rather an abstract ability that indirectly affects
performance in many tasks, then perhaps it is most concretely
described as the ability to do the kind of abstract problems that
appear in intelligence tests?  In which case, Boring's definition is
not as facetious as it first appears.

This definition also highlights the fact that the concept of
intelligence, and how it is measured, are intimately related.  In the
context of this paper we refer to these as definitions of
intelligence, and tests of intelligence, respectively.

\section{A Definition of Machine Intelligence}\label{sec:fmi}

\begin{quote}\it
``Indeed the guiding inspiration of cognitive science is that at a
  suitable level of abstraction, a theory of natural intelligence
  should have the same basic form as the theories that explain
  sophisticated computer systems.''
  \par\hfill --- {\sl J. Haugeland~\cite{Haugeland:81}}
\end{quote}

Having presented a very general informal definition of intelligence in
Subsection~\ref{subsec:idi}, we will now proceed to formalise this
definition mathematically in a way that is appropriate for machines.
We will then study some of the properties of this definition in the
remainder of this section.

\subsection{Basic agent-environment framework}
\label{subsec:aef}

Consider again our informal definition of intelligence:
\begin{quote}
\emph{Intelligence measures an agent's ability to achieve goals in a
  wide range of environments.}
\end{quote}

This definition contains three essential components: An agent,
environments and goals.  Clearly, the agent and the environment must
be able to interact with each other, specifically, the agent needs to
be able to send signals to the environment and also receive signals
being sent from the environment.  Similarly, the environment must be
able to receive and send signals to the agent.  In our terminology we
will adopt the agent's perspective on these communications and refer
to the signals sent from the agent to the environment as
\emph{actions}, and the signals sent from the environment to the agent
as \emph{perceptions}.

Our definition of an agent's intelligence also requires there to be
some kind of goal for the agent to try to achieve.  Perhaps an agent
could be intelligent, in an abstract sense, without having any
objective to apply its intelligence to.  Or perhaps the agent has no
desire to exercise its intelligence in a way that effects its
environment.  In either case, the agent's intelligence would be
unobservable and, more importantly, of no practical consequence.
Intelligence then, at least the concrete kind that interests us, comes
into effect when the agent has an objective that it actively pursues
by interacting with its environment.  Here we will refer to this
objective as its \emph{goal}.

The existence of a goal raises the problem of how the agent knows what
the goal is.  One possibility would be for the goal to be known in
advance and for this knowledge to be built into the agent.  The
problem with this however is that it limits each agent to just one
goal.  We need to allow agents that are more flexible, specifically,
we need to be able to inform the agent of what the goal is.  For
humans this is easily done using language.  In general however, the
possession of a sufficiently high level of language is too strong an
assumption to make about the agent.  Indeed, even for something as
intelligent as a dog or a cat, direct explanation is not very
effective.

Fortunately there is another possibility which is, in some sense, a
blend of the above two.  We define an additional communication channel
with the simplest possible semantics: A signal that indicates how good
the agent's current situation is.  We will call this signal the
\emph{reward}.  The agent's goal is then simply to maximise the amount
of reward it receives.  So in a sense its goal is fixed.  This is not
particularly limiting however, as we have not said anything about what
causes different levels of reward to occur.  In a complex setting the
agent might be rewarded for winning a game or solving a puzzle.  If
the agent is to succeed in its environment, that is, receive a lot of
reward, it must learn about the structure of the environment and in
particular what it needs to do in order to get reward.  Thus from a
broad perspective, the goal is flexible.  Not surprisingly, this is
exactly the way in which we condition an animal to achieve a goal: By
selectively rewarding certain behaviours (see
Subsection~\ref{subsec:animal}).  In a narrow sense the animal's goal
is fixed, perhaps to get more treats to eat, but in a broader sense it
is flexible as it may require doing a trick or solving a puzzle of our
choosing.

\begin{figure}[t]
\centerline{\includegraphics[width=0.45\columnwidth]{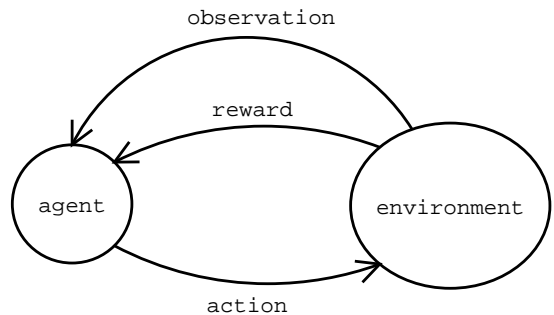}}
\caption{\label{agent-env}The agent and the environment interact by
sending action, observation and reward signals to each other.}
\end{figure}

In our framework we will include the reward signal as a part of the
perception generated by the environment.  The perceptions also
contain a non-reward part, which we will refer to as
\emph{observations}.  This now gives us the complete system of
interacting agent and environment, as illustrated in
Figure~\ref{agent-env}.  The goal, in the broad flexible sense, is
implicitly defined by the environment as this is what defines when
rewards are generated.  Thus, in the framework as we have defined it,
to test an agent in any given way it is sufficient to fully define the
environment.

This widely used and very flexible structure is in itself nothing new.
In artificial intelligence it is the framework used in reinforcement
learning \cite{Sutton:98}.  By appropriately renaming things, it also
describes the controller-plant framework used in control theory
\cite{Bertsekas:96}.  The interesting point for us is that this setup
follows naturally from our informal definition of intelligence and our
desire to keep things as general as possible.  The only difficulty was
how to deal with the notion of success, or profit.  This required the
existence of some kind of an objective or goal.  The most flexible and
elegant way to bring this into the framework was to use a simple
reward signal.

\begin{Example} \label{ex:twocoins1}
To make this model more concrete, consider the following ``Two Coins
Game''.  In each cycle two 50\textcent{} coins are tossed.  Before the
coins settle the player must guess at the number of heads that will
result: either 0, 1, or 2.  If the guess is correct the player gets to
keep both coins and then two new coins are produced and the game
repeats.  If the guess is incorrect the player does not receive any
coins, and the game is repeated.

In terms of the agent-environment model, the player is the agent and
the system that produces all the coins, tosses them and distributes
the reward when appropriate, is the environment.  The agent's actions
are its guesses at the number of heads in each iteration of the game:
0, 1 or 2.  The observation is the state of the coins when they
settle, and the reward is either \$0 or \$1.

It is easy to see that for unbiased coins the most likely outcome is 1
head and thus the optimal strategy for the agent is to always guess 1.
However if the coins are significantly biased it might be optimal to
guess either 0 or 2 heads depending on the bias.  If this were the
case, then after a number of iterations of the game an intelligent
agent would realise that the coins were probably biased and change its
strategy accordingly.
\end{Example}

With a little imagination, seemingly any sort of game, challenge,
problem or test can be expressed in this simple framework without too
much effort.  It should also be emphasised that this agent-environment
framework says nothing about how the agent or the environment actually
work; it only describes their roles.

\subsection{Formal agent-environment framework}
\label{subsec:formframe}

Having introduced the agent-environment framework, we will now
formalise it, along with the other components of our informal
definition of intelligence.  We begin with agent-environment
interaction.

\paragraph{Agent-environment interaction.}
The agent sends information to the environment by sending
\emph{symbols} from some finite set, for example, $\AA := \{
\mathtt{left}, \mathtt{right}, \mathtt{up}, \mathtt{down} \}$.  We
will call this set the \emph{action space} and denote it by $\AA$.
Similarly, the environment sends signals to the agent with symbols
from a finite set called the \emph{perception space}, which we will
denote $\PP$.  The \emph{reward space}, denoted by $\RR$, will always
be a subset of the rational unit interval $[0,1] \cap \QQQ$.  Every
perception consists of two separate parts; an observation and a
reward.  For example, we might have $\PP := \{ (\mathtt{cold}, 0.0),
(\mathtt{warm}, 1.0), (\mathtt{hot}, 0.3) \}$ where the first part
describes what the agent observes (cold, warm or hot) and the second
part describes the reward (0.0, 1.0, or 0.3).

To denote symbols being sent we will use the lower case variable names
$a$, $o$ and $r$ for actions, observations and rewards respectively.
We will also index these in the order in which they occur, thus $a_1$
is the agent's first action, $a_2$ is the second action and so on.
The agent and the environment will take turns at sending symbols,
starting with the environment.  This produces a history of
observations, rewards and actions which we will denote by, $o_1 r_1
a_1 o_2 r_2 a_2 o_3 r_3 a_3 o_4 \ldots$.  This turn taking behaviour
is not a serious restriction, nor is the fact that the first signal
sent is a perception.

\paragraph{The agent.}
Formally, the agent is a function, denoted by $\pi$, which takes the
current history as input and chooses the next action as output.  We do
not want to restrict the agent in any way, in particular we do not
require that it is deterministic.  A convenient way of representing
the agent then is as a probability measure over actions conditioned on
the complete interaction history.  Thus, $\pi( a_3 | o_1 r_1 a_1 o_2
r_2 )$ is the probability of action $a_3$ in the third cycle, given
that the current history is $o_1 r_1 a_1 o_2 r_2$.  A deterministic
agent is simply one that always assigns a probability of 1 to a single
action for any given history.  As the history that the agent can use
to select its action expands indefinitely, the agent need not be
Markovian.  Indeed, how the agent produces its distribution over
actions for any given history is left open.  In artificial
intelligence the agent will of course be a machine and so $\pi$ will
be a computable function.  In general however, $\pi$ could be
anything: An algorithm that generates the digits of $\sqrt{e}$ as
outputs, an incomputable function, or even a human pushing buttons on
a keyboard.

\paragraph{The environment.}
We define the environment, denoted by $\mu$, in a similar way.
Specifically, for any $k \in \NNN$ the probability of $o_k r_k$, given
the current interaction history $o_1 r_1 a_1 o_2 r_2 a_2 \ldots
o_{k-1} r_{k-1} a_{k-1}$, is given by the probability measure $\mu(
o_k r_k | o_1 r_1 a_1 o_2 r_2 a_2 \ldots o_{k-1} r_{k-1} a_{k-1} )$.
For the moment we will not place any further restrictions on the
environment.

\begin{Example}
To illustrate this formalism, consider again the Two Coins Game
introduced in Example~\ref{ex:twocoins1}.  Let $\PP := \{ 0, 1, 2 \}
\times \{ 0, 1 \}$ be the perception space representing the number of
heads after tossing the two coins and the value of the received
reward.  Likewise let $\AA := \{ 0, 1, 2 \}$ be the action space
representing the agent's guess at the number of heads that will occur.
Assuming two fair coins, we can represent this environment by $\mu$:
\begin{displaymath}
\mu( o_k r_k | o_1 \ldots a_{k-1}  ) :=  \left\{
\begin{array}{ll}
 {\frac{1}{4}}_{\phantom{|_{|_|}}} & \mathrm{if\ }o_k   =  a_{k-1} \in \{0,2\} \land r_k =1, \\
 {\frac{3}{4}}_{\phantom{|_{|_|}}} & \mathrm{if\ }o_k \neq a_{k-1} \in \{0,2\} \land r_k =0, \\
 {\frac{1}{2}}_{\phantom{|_{|_|}}} & \mathrm{if\ }o_k   =  a_{k-1} = 1 \land r_k=1, \\
 {\frac{1}{2}}_{\phantom{|_{|_|}}} & \mathrm{if\ }o_k \neq a_{k-1} = 1 \land r_k=0, \\
0 &  \!\mathrm{otherwise}.
\end{array} \right.
\end{displaymath}
An agent that performs well in this environment would be,
\begin{displaymath}
\pi( a_k | o_1 r_1 a_1 \ldots o_k r_k ) := \left\{
\begin{array}{ll}
1 & \mathrm{for\ } a_k = 1,\\
0 & \mathrm{otherwise}.
\end{array} \right.
\end{displaymath}
That is, always guess that
one head will be the result of the two coins being tossed.  A more
complex agent might keep count of how many heads occur in each cycle
and then adapt its strategy if it seems that the coins are
sufficiently biased.
\end{Example}

\paragraph{Measure of success.}
Our next task is to formalise the idea of ``profit'' or ``success''
for an agent.  Informally, we know that the agent must try to maximise
the amount of reward it receives, however this could mean several
different things.  For example, one agent might quickly find a way to
get a reward of 0.9 in every cycle.  After 100 cycles it will have
received a total reward of about 90 with an average reward per cycle
of close to 0.9.  A second agent might spend the first 80 cycles
exploring different actions and their consequences, during which time
its average reward might only be 0.2.  Having done this exploration
however, it might then know a way to get a reward of 1.0 in every
cycle.  Thus after 100 cycles its total reward is only $80 \times 0.2
+ 20 \time 1.0 = 36$, giving an average reward per cycle of just 0.36.
After 1,000 cycles however, the second agent will be performing much
better than the first.

Which agent is the better one?  The answer depends on how we value
reward in the near future versus reward in the more distant future.
In some situations we may want our agent to perform well fairly
quickly, in others we might only care that it eventually reaches a
level of performance that is as high as possible.

A standard way of formalising this is to scale the value of rewards so
that they decay geometrically into the future at a rate given by a
discount parameter $\gamma \in (0,1)$.  For example, with $\gamma =
0.95$ a reward of $0.7$ that is $10$ time steps into the future would
be given a value of $0.7 \times (0.95)^{10} \approx 0.42$.  At $100$
time steps into the future a reward of $0.7$ would have a value of
just over $0.004$.  By increasing $\gamma$ towards 1 we weight long
term rewards more heavily, conversely by reducing it we weight them
less so.  In other words, this parameter controls how short term
greedy, or long term farsighted, the agent should be.

To work out the expected future value for a given agent and
environment interacting, we take the sum of these discounted rewards
into the infinite future and work out its expected value,
\begin{equation}
\label{eqn:disval}
V^\pi_\mu (\gamma) := \: \frac{1}{\Gamma} \E \left( \sum_{i=1}^\infty
\gamma^i r_i \right).
\end{equation}
In the above, $r_i$ is the reward in cycle $i$ of a given history,
$\gamma$ is the discount rate, $\gamma^i$ is the discount applied to
the $i^{th}$ reward into the future, the normalising constant is
$\Gamma := \sum_{i = 1}^\infty \gamma^i$, and the expected value is
taken over all possible interaction sequences between the agent $\pi$
and the environment $\mu$.

Under geometric discounting an agent with $\gamma=0.95$ will not plan
further than about 20 cycles ahead.  Thus we say that the agent has a
constant effective horizon of ${1\over 1-\gamma}$.  Since we are
interested in universal intelligence, a limited farsightedness is not
acceptable because for every horizon there is a task that needs a
larger horizon to be solved.  For instance, while a horizon of 5 is
sufficient for tic-tac-toe, it is insufficient for chess.  Clearly,
geometric discounting has not solved the problem of how to weight near
term rewards versus long term rewards, it has simply expressed this
weighting as a parameter.  What we require is a single definition of
machine intelligence, not a range of definitions that vary according
to a free parameter.

A more promising candidate for universal discounting is the
near-harmonic, or quadratic discount, where we replace $\gamma^i$ in
Equation~\ref{eqn:disval} by $1/i^{2}$ and modifying $\Gamma$
accordingly.  This has some interesting properties, in particular the
agent needs to look forward into the future in a way that is
proportional to its current age.  This is appealing since it seems
that humans of age $k$ years usually do not plan their lives for more
than, perhaps, the next $k$ years.  More importantly, it allows us to
avoid the problem of having to choose a global time scale or effective
horizon~\cite{Hutter:04uaibook}.

Although harmonic discounting has a number of attractive
properties~\cite{Hutter:06discount}, an even simpler and more general
solution is possible.  If we look at the value function in
Equation~\ref{eqn:disval}, we see that discounting plays two roles.
Firstly, it normalises rewards received so that their sum is always
finite.  Secondly, it weights the reward at different points in the
future which in effect defines a temporal preference.  A direct way to
solve both of these problems, without needing an external parameter,
is to simply require that the total reward returned by the environment
can never exceed 1.  For such a reward summable environment $\mu$, it
follows that the expected value of the sum of rewards is also finite
and thus discounting is no longer required,
\begin{equation}\label{eqn:unival}
V^{\pi}_{\mu} := \: \E \left( \sum_{i=1}^\infty r_i \right)
\leq 1.
\end{equation}

One way of viewing this is that the rewards returned by the
environment now have the temporal preference already factored in.  The
cost is that this is an additional condition that we place on the
space of environments.  Previously we required that each reward signal
was in a subset of $[0,1] \cap \QQQ$, now we have the additional
constraint that the reward sum is always bounded (see
Subsection~\ref{subsec:crit} for further discussion about why this
constraint is reasonable).

\paragraph{Space of environments.}
Intelligence is not simply the ability to perform well at a narrowly
defined task; it is much broader.  An intelligent agent is able to
adapt and learn to deal with many different situations, kinds of
problems and types of environments.  In our informal definition this
was described as the agent's general ability to perform well in a
``wide range of environments.''  This flexibility is a defining
characteristic and one of the most important differences between
humans and many current AI systems: While Gary Kasparov would still be
a formidable player if we were to change the rules of chess, IBM's
Deep Blue chess super computer would be rendered useless without
significant human intervention.

As our goal is to produce a definition of intelligence that is as
broad and encompassing as possible, the space of environments used in
our definition should be as large as possible.  As the environment is
a probability measure with a certain structure, an obvious possibility
would be to consider the space of all probability measures of this
form.  Unfortunately, this extremely broad class of environments
causes serious problems.  As the space of all probability measures is
uncountably infinite, some environments cannot be described in a
finite way and so are incomputable.  This would make it impossible, by
definition, to test an agent in such an environment using a computer.
Further, most environments would be infinitely complex and have little
structure for the agent to learn from.

The solution then, is to require the environmental probability
measures to be computable.  Not only is this condition necessary if we
are to have an effective measure of intelligence, it is also not as
restrictive as it might first appear.  There are still an infinite
number of environments with no upper bound on their maximal
complexity.  Also, it is only the measure that describes the
environment that is computable, and so the way in which the
environment responds does not have to be deterministic.  For example,
although a typical sequence of 1's and 0's generated at random by
flipping a coin is not computable, the probability measure that
describes this distribution is computable and thus it is included in
our space of possible environments.  Indeed there is currently no
evidence that the physical universe cannot be simulated by a Turing
machine in the above sense (for further discussion of this point see
Subsection~\ref{subsec:crit}).  This appears to be the largest
reasonable space of environments.

\subsection{A formal definition of machine intelligence}
\label{subsec:fmi}

In order to define an overall measure of performance, we need to find
a way to combine an agent's performance in many different environments
into a single overall measure.  As there are an infinite number of
environments, we cannot simply take a uniform distribution over them.
Mathematically, we must weight some environments higher than others.
But how?

Consider the agent's perspective on this situation: There exists a
probability measure that describes the true environment, however this
measure is not known to the agent.  The only information the agent has
are some past observations of the environment.  From these, the agent
can construct a list of probability measures that are consistent with
the observations.  We call these potential explanations of the true
environment, hypotheses.  As the number of observations increases, the
set of hypotheses shrinks and hopefully the remaining hypotheses
become increasingly accurate at modelling the true environment.

The problem is that in any given situation there will likely be a
large number of hypotheses that are consistent with the current set of
observations.  Thus, if the agent is going to predict which hypotheses
are the most likely to be correct, it must resort to something other
than just the observational information that it has.  This is a
frequently occurring problem in inductive inference for which the most
common approach is to invoke the principle of Occam's razor:
\begin{quote}
\emph{Given multiple hypotheses that are consistent with the data, the
  simplest should be preferred.}
\end{quote}
This is generally considered the rational and intelligent thing to do
\cite{Wallace:05}, indeed IQ tests often implicitly test an
individual's ability to use Occam's razor, as pointed out in
Subsection~\ref{subsec:sti}.

\begin{Example}
Consider the following type of question which commonly appears in
intelligence tests.  There is a sequence such as 2, 4, 6, 8, and the
test subject needs to predict the next number.  Of course the pattern
is immediately clear: The numbers are increasing by 2 each time, or
more mathematically, the $k^{th}$ item is given by $2k$.  An
intelligent person would easily identify this pattern and predict the
next digit to be 10.  However, the polynomial $2k^4 -20k^3 +70k^2 -98k
+48$ is also consistent with the data, in which case the next number
in the sequence would be 58.  Why then, even if we are aware of the
larger polynomial, do we consider the first answer to be the most
likely one?  It is because we apply, perhaps unconsciously, the
principle of Occam's razor.  The fact that intelligence tests
define this as the ``correct'' answer, shows us that using Occam's
razor is considered the intelligent thing to do.  Thus, although we do
not usually mention Occam's razor when defining intelligence, the
ability to effectively use it is an important facet of intelligent
behaviour.
\end{Example}

In some cases we may even consider the correct use of Occam's razor to
be a more important demonstration of intelligence than achieving a
successful outcome.  Consider, for example, the following game:

\begin{Example}
A questioner lays twenty \$10 notes out on a table before you and then
points to the first one and asks ``Yes or No?''.  If you answer
``Yes'' he hands you the money.  If you answer ``No'' he takes it from
the table and puts it in his pocket.  He then points to the next \$10
note on the table and asks the same question.  Although you, as an
intelligent agent, might experiment with answering both ``Yes'' and
``No'' a few times, by the $13^{th}$ round you would have decided that
the best choice seems to be ``Yes'' each time.  However what you do
not know is that if you answer ``Yes'' in the $13^{th}$ round then the
questioner will pull out a gun and shoot you!  Thus, although
answering ``Yes'' in the $13^{th}$ round is the most intelligent
choice, given what you know, it is not the most successful one.  An
exceptionally dim individual may have failed to notice the obvious
relationship between answers and getting the money, and thus might
answer ``No'' in the $13^{th}$ round, thereby saving his life due to
what could truly be called ``dumb luck''.
\end{Example}

What is important then, is not that an intelligent agent succeeds in
any given situation, but rather that it takes actions that we would
expect to be the most likely ones to lead to success.  Given adequate
experience this might be clear, however often experience is not
sufficient and one must fall back on good prior assumptions about the
world, such as Occam's razor.  It is important then that we test the
agents in such a way that they are, at least on average, rewarded for
correctly applying Occam's razor, even if in some cases this leads
to failure.

There is another subtlety that needs to be pointed out.  Often
intelligence is thought of as the ability to deal with complexity.  Or
in the words of the psychologist Gottfredson, ``\ldots $g$ is the
ability to deal with cognitive complexity --- in particular, with
complex information processing.''\cite{Gottfredson:97} It is tempting
then to equate the difficultly of an environment with its complexity.
Unfortunately, things are not so straightforward.  Consider the
following environment:
\begin{Example}
Imagine a very complex environment with a rich set of relationships
between the agent's actions and observations.  The measure that
describes this will have a high complexity.  However, also imagine
that the reward signal is always maximal no matter what the agent
does.  Thus, although this is a very complex environment in which the
agent is unlikely to be able predict what it will observe next, it is
also an easy environment in the sense that all policies are optimal,
even very simple ones that do nothing at all.  The environment
contains a lot of structure that is irrelevant to the goal that the
agent is trying to achieve.
\end{Example}

From this perspective, a problem is thought of as being difficult if
the simplest good solution to the problem is complex.  Easy problems
on the other hand are those that have simple solutions.  This is a
very natural way to think about the difficulty of problems, or in our
terminology, environments.

Fortunately, this distinction does not affect our use of Occam's
razor.  When we talk about an hypothesis, what we mean is a potential
model of the environment from the agent's perspective, not just a
model that is sufficient with respect to the agent's goal.  From the
agent's perspective, an incorrect hypothesis that fails to model much
of the environment may be optimal if the parts of the environment that
the hypothesis fails to model are not relevant to receiving reward.
However, when Occam's razor is applied, we apply it with respect to
the complexity of the hypotheses, not the complexity of good solutions
with respect to an objective.  Thus, to reward agents on average for
correctly using Occam's razor, we must weight the environments
according to their complexity, not their difficulty.

Our remaining problem now is to measure the complexity of
environments.  The Kolmogorov complexity of a binary string $x$ is
defined as being the length of the shortest program that
computes $x$:
\[
  K(x) \;:=\; \min_p\{l(p): \UU(p)=x\},
\]
where $p$ is a binary string which we call a \emph{program}, $l(p)$ is
the length of this string in bits, and $\UU$ is a prefix universal
Turing machine $\UU$ called the \emph{reference machine}.

To gain an intuition for how this works, consider a binary string
$0000\ldots0$ that consists of a trillion $0$s.  Although this string
is very long, it clearly has a simple structure and thus we would
expect it to have a low complexity.  Indeed this is the case because
we can write a very short program $p$ that simply loops a trillion
times outputting a $0$ each time.  Similarity, other strings with
simple patterns have a low Kolmogorov complexity.  On the other hand,
if we consider a long irregular random string $111010110000010 \ldots
$ then it is much more difficult to find a short program that outputs
this string.  Indeed it is possible to prove that there are so many
strings of this form, relative to the number of short programs, that
in general it is impossible for long random strings to have short
programs.  In other words, they have high Kolmogorov complexity.

An important property of $K$ is that it is nearly independent of the
choice of $\UU$.  To see why, consider what happens if we switch from
$\UU$, in the above definition of $K$, to some other universal Turing
machine $\UU'$.  Due to the universality property of $\UU'$, there
exists a program $q$ that allows $\UU'$ to simulate $\UU$.  Thus, if
we give $\UU'$ both $q$ and $p$ as inputs, it can simulate $\UU$
running $p$ and thereby compute $\UU(p)$.  It follows then that
switching from $\UU$ to $\UU'$ in our definition of $K$ above incurs
at \emph{most} an additional cost of $l(q)$ bits in minimal program
length.  The constant $l(q)$ is independent of which string $x$ we are
measuring the complexity of, and for reasonable universal Turing
machines, this constant will be small.  This invariance property makes
$K$ an excellent universal complexity measure.  For an extensive
treatment of Kolmogorov complexity see \cite{Li:97} or
\cite{Calude:02}.

In our current application we need to measure the complexity of the
computable measures that describe environments.  It can be shown that
this set can be enumerated $\mu_1, \mu_2, \mu_3, \ldots$ (see
Theorem~4.3.1 in \cite{Li:97}).  Using a simple encoding method we can
express each index as a binary string, written $\langle i \rangle$.
In a sense this binary string is a description of an environment with
respect to our enumeration.  This lets us define the complexity of an
environment $\mu_i$ to be $K(\mu_i) := K(\langle i \rangle )$.
Intuitively, if a short program can be used to describe the program
for an environment $\mu_i$, then this environment will have a low
complexity.

This answers our problem of needing to be able to measure the
complexity of environments, but we are not done yet.  In order to
formalise Occam's razor we need to have a way to assign an a priori
probability to environments in such a way that complex environments
are less likely, and simple environments more likely.  If we consider
that each environment $\mu_i$ is described by a minimal length program
that is a binary string, then the natural way to do this is to
consider each additional bit of program length to reduce the
environment's probability by one half, reflecting the fact that each
bit has two possible states.  This gives us what is known as the
\emph{algorithmic probability distribution} over the space of
environments, defined $2^{-K(\mu)}$.  This distribution has powerful
properties that essentially solve long-standing open philosophical,
statistical, and computational problems in the area of inductive
inference~\cite{Hutter:07uspx}.  Furthermore, the distribution can be
used to define powerful universal learning agents that have provably
optimal performance~\cite{Hutter:04uaibook}.

Bringing all these pieces together, we can now define our formal
measure of intelligence for arbitrary systems.  Let $E$ be the space
of all computable reward summable environmental measures with respect
to the reference machine $\UU$, and let $K$ be the Kolmogorov
complexity function.  The expected performance of agent $\pi$ with
respect to the universal distribution $2^{-K(\mu)}$ over the space of
all environments $E$ is given by,
\[
\Upsilon(\pi) \; := \; \sum_{\mu \in E} \, 2^{-K(\mu)} \, V^\pi_\mu.
\]
We call this the \emph{universal intelligence} of agent $\pi$.

Consider how this equation corresponds to our informal definition.  We
needed a measure of an agent's general ability to achieve goals in a
wide range of environments.  Clearly present in the equation is the
agent $\pi$, the environment $\mu$ and, implicit in the environment, a
goal.  The agent's ``ability to achieve'' is represented by the value
function $V^\pi_\mu$.  By a ``wide range of environments'' we have
taken the space of all well defined reward summable environments,
where these environments have been characterised as computable
measures in the set $E$.  Occam's razor is given by the term
$2^{-K(\mu)}$ which weights the agent's performance in each
environment inversely proportional to its complexity.  The definition
is very general in terms of which sensors or actuators the agent might
have as all information exchanged between the agent and the
environment takes place over very general communication channels.
Finally, the formal definition places no limits on the internal
workings of the agent.  Thus, we can apply the definition to any
system that is able to receive and generate information with view to
achieving goals.  The main drawback, however, is that the Kolmogorov
complexity function $K$ is not computable and can only be
approximated.  This is an important point that we will return to
later.

\subsection{Universal intelligence of various agents}
\label{subsec:eaae}

In order to gain some intuition for our definition of universal
intelligence, in this subsection we will consider a range of different
agents and their relative degrees of universal intelligence.

\paragraph{A random agent.}
The agent with the lowest intelligence, at least among those that are
not actively trying to perform badly, would be one that makes
uniformly random actions.  We will call this $\pi^\mathtt{rand}$.
Although this is clearly a weak agent, we cannot simply conclude that
the value of $V^{\pi^\mathtt{rand}}_\mu$ will always be low as some
environments will generate high reward no matter what the agent does.
Nevertheless, in general such an agent will not be very successful as
it will fail to exploit any regularities in the environment, no matter
how simple they are.  It follows then that the values of
$V^{\pi^\mathtt{rand}}_\mu$ will typically be low compared to other
agents, and thus $\Upsilon (\pi^\mathtt{rand})$ will be low.
Conversely, if $\Upsilon (\pi^\mathtt{rand})$ is very low, then the
equation for $\Upsilon$ implies that for simple environments, and many
complex environments, the value of $V^{\pi^\mathtt{rand}}_\mu$ must
also be relatively low.  This kind of poor performance in general is
what we would expect of an unintelligent agent.

\paragraph{A very specialised agent.}
From the equation for $\Upsilon$, we see that an agent could have very
low universal intelligence but still perform extremely well at a few
very specific and complex tasks.  Consider, for example, IBM's Deep
Blue chess supercomputer, which we will represent by
$\pi^\mathtt{dblue}$.  When $\mu^\mathtt{chess}$ describes the game of
chess, $V^{\pi^\mathtt{dblue}}_{\mu^\mathtt{chess}}$ is very high.
However $2^{-K(\mu^\mathtt{chess})}$ is small, and for $\mu \neq
\mu^\mathtt{chess}$ the value function will be low as
$\pi^\mathtt{dblue}$ only plays chess.  Therefore, the value of
$\Upsilon (\pi^\mathtt{dblue})$ will be very low.  Intuitively, this
is because Deep Blue is too inflexible and narrow to have general
intelligence; a characteristic weakness of specialised artificial
intelligence systems.

\paragraph{A general but simple agent.}
Imagine an agent that performs very basic learning by building up a table
of observation and action pairs and keeping statistics on the rewards
that follow.  Each time an observation that it has been seen before
occurs, the agent takes the action with highest estimated expected
reward in the next cycle with 90\% probability, or a random action
with 10\% probability.  We will call this agent $\pi^\mathtt{basic}$.
It is immediately clear that many environments, both complex and very
simple, will have at least some structure that such an agent would
take advantage of.  Thus, for almost all $\mu$ we will have
$V^{\pi^\mathtt{basic}}_\mu > V^{\pi^\mathtt{rand}}_\mu$ and so
$\Upsilon (\pi^\mathtt{basic}) > \Upsilon (\pi^\mathtt{rand})$.
Intuitively, this is what we would expect as $\pi^\mathtt{basic}$,
while very simplistic, is surely more intelligent than
$\pi^\mathtt{rand}$.

Similarly, as $\pi^\mathtt{dblue}$ will fail to take advantage of even
trivial regularities in some of the most basic environments, $\Upsilon
(\pi^\mathtt{basic}) > \Upsilon (\pi^\mathtt{dblue})$.  This is
reasonable as our aim is to measure a machine's level of general
intelligence.  Thus an agent that can take advantage of basic
regularities in a wide range of environments should rate more highly
than a specialised machine that fails outside of a very limited
domain.

\paragraph{A simple agent with more history.}
The first order structure of $\pi^\mathtt{basic}$, while very general,
will miss many simple exploitable regularities.  Consider the
following environment $\mu^\mathtt{alt}$.  Let $\RR = [0,1] \cap
\QQQ$, $\AA = \{ \mathtt{up}, \mathtt{down} \}$ and $\OO = \{\eps\}$,
where $\eps$ is the empty string.  In cycle $k$ the environment
generates a reward of $2^{-k}$ each time the agent's action is
different to its previous action.  Otherwise the reward is 0.  We can
define this environment formally,
\begin{displaymath}
\mu^\mathtt{alt}( o_k r_k | o_1 \ldots a_{k-1} ) := \left\{
\begin{array}{ll}
 1 & \mathrm{if\ }a_{k-1} \neq a_{k-2} \land r_k =2^{-k}, \\
 1 & \mathrm{if\ }a_{k-1}   =  a_{k-2} \land r_k =0, \\
 0 &  \!\mathrm{otherwise}.
\end{array} \right.
\end{displaymath}
Clearly the optimal strategy for an agent is simply to alternate
between the actions $\mathtt{up}$ and $\mathtt{down}$.  Even though
this is very simple, this strategy requires the agent to correlate its
current action with its previous action, something that
$\pi^\mathtt{basic}$ cannot do.

A natural extension of $\pi^\mathtt{basic}$ is to use a longer history
of actions, observations and rewards in its internal table.  Let
$\pi^\mathtt{2back}$ be the agent that builds a table of statistics
for the expected reward conditioned on the last two actions, rewards
and observations.  It is immediately clear that $\pi^\mathtt{2back}$
will exploit the structure of the $\mu^\mathtt{alt}$ environment.
Furthermore, by definition $\pi^\mathtt{2back}$ is a generalisation of
$\pi^\mathtt{basic}$ and thus it will adapt to any regularity that
$\pi^\mathtt{basic}$ can adapt to.  It follows then that in general
$V^{\pi^\mathtt{2back}}_\mu > V^{\pi^\mathtt{basic}}_\mu$ and so
$\Upsilon (\pi^\mathtt{2back}) > \Upsilon (\pi^\mathtt{basic})$, as we
would intuitively expect.  In the same way we can extend the history
that the agent utilises back further and produce even more powerful
agents that are able to adapt to more lengthy temporal structures and
which will have still higher machine intelligence.

\begin{figure}[t]
\centerline{\includegraphics[width=0.65\columnwidth]{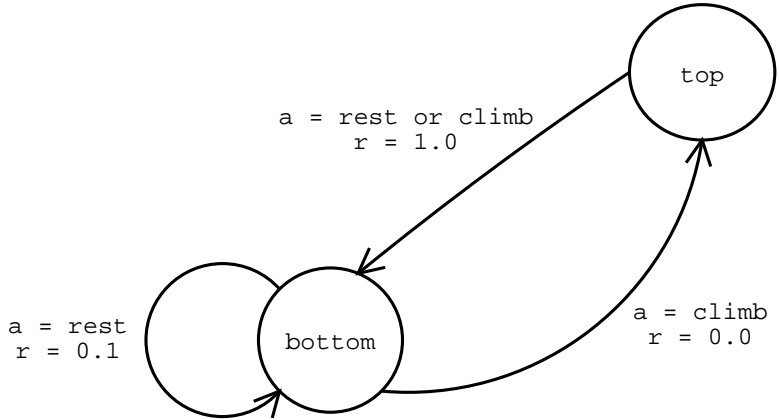}}
\caption{\label{slide}A simple game in which the agent climbs a
  playground slide and slides back down again.  A shortsighted agent
  will always just rest at the bottom of the slide.}
\end{figure}

\paragraph{A simple forward looking agent.}
In some environments simply trying to maximise the next reward is not
sufficient, the agent must also take into account the rewards that are
likely to follow further into the future, that is, the agent must plan
ahead.  Consider the following environment $\mu^\mathtt{slide}$.  Let
$\RR = [0,1] \cap \QQQ$, $\AA = \{ \mathtt{rest}, \mathtt{climb} \}$
and $\OO = \{\eps\}$.  Intuitively, there is a slide such as you would
see in a playground.  The agent can rest at the bottom of the slide,
for which it receives a reward of $2^{-k-4}$.  The alternative is to
climb the slide, which gives a reward of 0.  Once at the top of the
slide the agent always slides back down no matter what action is
taken; this gives a reward of $2^{-k}$.  This is illustrated in
Figure~\ref{slide}.  The environment is completely deterministic.

Because climbing receives a reward of 0, while resting receives a
reward of $2^{-k-4}$, a very shortsighted agent that only tries to
maximise the reward in the next cycle will choose to stay at the
bottom of the slide.  Both $\pi^\mathtt{basic}$ and
$\pi^\mathtt{2back}$ have this problem; though they also take random
actions 10\% probability and so will occasionally climb the slide by
chance.  Clearly this is not optimal in terms of total reward over
time.

We can extend the $\pi^\mathtt{2back}$ agent again by defining a new
agent $\pi^\mathtt{2forward}$ that with 90\% probability chooses its
next action to maximise not just the next reward, but $\hat{r}_{k+1} +
\hat{r}_{k+2}$, where $\hat{r}_{k+1}$ and $\hat{r}_{k+2}$ are the
agent's estimates of the next two rewards.  As the estimate of
$\hat{r}_{k+2}$ will potentially depend not only on $a_k$, but also on
$a_{k+1}$, the agent assumes that $a_{k+1}$ is chosen to simply
maximise the estimated reward $\hat{r}_{k+2}$.

The $\pi^\mathtt{2back}$ agent can see that by missing out on the
resting reward of $2^{-k-4}$ for one cycle and climbing, a
greater reward of $2^{-k}$ will be had when sliding back down the
slide in the following cycle.

By definition $\pi^\mathtt{2forward}$ generalises $\pi^\mathtt{2back}$
in a way that more closely reflects the value function $V$ and thus in
general $V^{\pi^\mathtt{2forward}}_\mu > V^{\pi^\mathtt{2back}}_\mu$.
It then follows that $\Upsilon (\pi^\mathtt{2forward}) > \Upsilon
(\pi^\mathtt{2back})$ as we would intuitively expect for this more
powerful agent.

\vspace{1em}

In a similar way agents of increasing complexity and adaptability can
be defined which will have still greater intelligence.  However with
more complex agents it is usually difficult to theoretically establish
whether one agent has more or less universal intelligence than
another.  Nevertheless, in the simple examples above we saw that the
more flexible and powerful an agent was, the higher its universal
intelligence.

\paragraph{A very intelligent agent.}
A very smart agent would perform well in simple environments, and
reasonably well compared to most other agents in more complex
environments.  From the equation for universal intelligence this would
clearly produce a very high value for $\Upsilon$.  Conversely, if
$\Upsilon$ was very high then the equation for $\Upsilon$ implies that
the agent must perform well in most simple environments and reasonably
well in many complex ones also.

\paragraph{A super intelligent agent.}
Consider what would be required to maximise the value of $\Upsilon$.
By definition, a ``perfect'' agent would always pick the action which
had greatest expected future reward.  To do this, for every
environment $\mu \in E$ the agent must take into account how likely it
is that it is facing $\mu$ given the interaction history so far, and
the prior probability of $\mu$, that is, $2^{-K(\mu)}$.  It would then
consider all possible future interactions that might occur, and how
likely they are, and from this select the action in the current cycle
that maximises the expected future reward.

This perfect theoretical agent is known as AIXI.  It has been
precisely defined and studied at length in \cite{Hutter:04uaibook}
(see \cite{Hutter:07aixigentle} for a shorter exposition).  The
connection between universal intelligence and AIXI is not
coincidental: $\Upsilon$ was originally derived from the so called
``intelligence order relation'' (see Definition 5.14 in
\cite{Hutter:04uaibook}), which in turn was constructed to reflect the
equations for AIXI.  As such we can define the upper bound on
universal intelligence to be,
\[
\bar{\Upsilon} \: := \: \max_\pi \Upsilon( \pi ) \: = \: \Upsilon
\left( \pi^{AIXI} \right).
\]

AIXI is not computable due to the incomputability of $K$, and even if
$K$ were computable, accurately computing the expectations to maximise
future expected rewards would be practically infeasible.
Nevertheless, AIXI is interesting from a theoretical perspective as it
defines, in an elegant way, what might be considered to be the perfect
theoretical artificial intelligence.  Indeed many strong optimality
properties have been proven for AIXI.  For example, it has been proven
that AIXI converges to optimal performance in any environment where
this is at all possible for a general agent (see Theorem 5.34 of
\cite{Hutter:04uaibook}).  This optimality result includes ergodic
Markov decision processes, prediction problems, classification
problems, bandit problems and many
others~\cite{Legg:04env,Legg:04mdp}.  These mathematical results prove
that agents with very high universal intelligence are extremely
powerful and general.

\paragraph{A human.}
For extremely simple environments, a human should be able to identify
their simple structure and exploit this to maximise reward.  However,
for more complex environments it is hard to know how well a human
would perform.  Much of the human brain is set up to process
information from the human sense organs, and thus is quite
specialised.  Perhaps the amount of universal machine intelligence
that a human has is not that high compared to some machine learning
algorithms?  It is difficult to know without experimental results.

\subsection{Properties of universal intelligence}
\label{subsec:anal}

What we have presented is a definition of machine intelligence, it is
not a practical test of machine intelligence, indeed the value of
$\Upsilon$ is not computable due to the use of Kolmogorov complexity.
The difference between the definition of something and practical tests
is important to keep in mind.  In some cases tests are based on a
definition or theory of intelligence.  In other cases, as we will see
in the next section, what is presented is some where between a fully
encompassing definition, and a realistically practical test.  Thus the
distinction between tests and definitions is not always clear.

Here our goal has simply been to define the concept of machine
intelligence in the most general, powerful and elegant way.  In future
research we will explore ways to approximate this ideal with a
practical test.  Naturally the process of estimation will introduce
weaknesses and flaws that the original definition did not have.  For
example, while the definition considers the general performance of an
agent over all computable environments with bounded reward sum, in
practice a test could only ever estimate this by testing the agent on
a finite sample of environments.

A similar situation arises when defining randomness for sequences.  In
essence, we consider an infinite sequence to be Martin-L\"{o}f random
when it has no significant regularity. This lack of regularity is
equivalent to saying that the sequence cannot be compressed in any
significant way, and thus we can characterise randomness using
Kolmogorov complexity.  Naturally, we cannot test a sequence for every
possible regularity, which is equivalent to saying that we cannot
compute its Kolmogorov complexity.  We can however test sequences for
randomness by checking them for a large number of statistical
regularities, indeed this is what is done in practice.  Of course,
just because a sequence passes all our tests does not mean that it
must be random.  There could always be some deeper structure to the
sequence that our tests were not able to detect.  All we can say is
that the sequence seems random with respect to our ability to detect
patterns.

Some might argue that the definition of something should not just
capture the concept, it should also be practical.  For example, the
definition of intelligence should be such that intelligence can be
easily measured.  The above example, however, illustrates why this
approach is sometimes flawed: If we were to \emph{define} randomness
with respect to a particular set of tests, then one could specifically
construct a sequence that followed a regular pattern in such a way
that it passed all of our randomness tests.  This would completely
undermine our definition of randomness.  A better approach is to
define the concept in the strongest and cleanest way possible, and
then to accept that our ability to test for this ideal has
limitations.  In other words, our task is to find better and more
effective tests, not to redefine what it is that we are testing for.
This is the attitude we have taken here, though in this paper our
focus is almost entirely on the first part, that is, establishing a
strong theoretical definition of machine intelligence.

Although some of the criteria by which we judge practical tests of
intelligence are not relevant to a pure definition of intelligence,
many of the desirable properties are similar.  Thus to understand the
strengths and weaknesses of our definition, consider again the
desirable properties for a test of intelligence from
Subsection~\ref{subsec:desirable}.

\paragraph{Valid.}
The most important property of any proposed formal definition of
intelligence is that it does indeed describe something that can
reasonably be called ``intelligence''.  Essentially, this is the core
argument of this report so far: We have taken a mainstream informal
definition and step by step formalised it.  Thus, so long as our
informal definition is reasonable, and our formalisation argument
holds, the result can reasonably be described as a formal definition
of intelligence.

\paragraph{Meaningful.}
As we saw in the previous section, universal intelligence orders the
power and adaptability of simple agents in a natural way.
Furthermore, a high value of $\Upsilon$ implies that the agent
performs well in most simple and moderately complex environments.
Such an agent would be an impressively powerful and flexible piece of
technology, with many potential uses.  Clearly then, universal
intelligence is inherently meaningful, independent of whether or not
one considers it to be a measure of intelligence.

\paragraph{Informative.}
$\Upsilon(\pi)$ is a real value that is independent of the performance
of other possible agents.  Thus we can make direct comparisons between
different agents on a single scale.  This property is important if we
want to use this measure to study new algorithms.

\paragraph{Wide range.}
As we saw in the previous section, universal intelligence is able to
order the intelligence of even the most basic agents such as
$\pi^\mathtt{rand}$, $\pi^\mathtt{basic}$, $\pi^\mathtt{2back}$ and
$\pi^\mathtt{2forward}$.  At the other extreme we have the theoretical
super intelligent agent AIXI which has maximal $\Upsilon$ value.
Thus, universal intelligence spans trivial learning algorithms right
up to super intelligent agents.  This seems to be the widest range
possible for a measure of machine intelligence.

\paragraph{General.}
As the agent's performance on all well defined environments is
factored into its $\Upsilon$ value, a broader performance metric is
difficult to imagine.  Indeed, a well defined measure of intelligence
that is broader than universal intelligence would seem to contradict
the Church-Turing thesis as it would imply that we could effectively
measure an agent's performance for some well defined problem that was
outside of the space of computable measures.

\paragraph{Unbiased.}
In a standard intelligence test, an individual's performance is judged
on specific kinds of problems, and then these scores are combined to
produce an overall result.  Thus a test's outcome is a product of
which types of problems it uses and how each score is weighted to
produce the end result.  Unfortunately, how we do this is a product of
many things, including our culture, values and the theoretical
perspective on intelligence that we have taken.  For example, while
one intelligence test could contain many logical puzzle problems,
another might be more linguistic in emphasis, while another stresses
visual reasoning.  Modern intelligence tests like the Stanford-Binet
try to minimise this problem by covering the most important areas of
human reasoning both verbally and non-verbally.  This helps but it is
still very anthropocentric as we are still only testing those
abilities that we think are important for human intelligence.

For an intelligence measure for arbitrary machines we have to base the
test on something more general and principled: Universal Turing
computation.  As all proposed models of computation have thus far been
equivalent in their expressive power, the concept of computation
appears to be a fundamental theoretical property rather than the
product of any specific culture.  Thus, by weighting different
environments depending on their Kolmogorov complexity, and considering
the space of all computable environments, we have avoided having to
define intelligence with respect to any particular culture, species
etc.

Unfortunately, we have not entirely removed the problem.  The
environmental distribution $2^{-K(\mu)}$ that we have used is
invariant, up to a multiplicative constant, to changes in the
reference machine $\UU$.  Although this affords us some protection,
the relative intelligence of agents can change if we change our
reference machine.  One approach to this problem is to limit the
complexity of the reference machine, for example by limiting its
state-symbol complexity.  We expect that for highly intelligent
machines that can deal with a wide range of environments of varying
complexity, the effect of changing from one simple reference machine
to another will be minor.  For simple agents, such as those considered
in Subsection~\ref{subsec:eaae}, the ordering of their machine
intelligence was also not particularly sensitive to natural choices of
reference machine.  Recently attempts have been made to make
algorithmic probability completely unique and objective by identifying
which universal Turing machines are, in some sense, the most
simple~\cite{Mueller:06}.  Unfortunately however, an elegant solution
to this problem has not yet been found.

\paragraph{Fundamental.}
Universal intelligence is based on computation, information and
complexity.  These are fundamental concepts that seem unlikely to
change in the future with changes in technology.  Indeed, if they were
to change, the implications would drastically affect the entire field
of computer science, not just this work.

\paragraph{Formal.}
Universal intelligence is expressed as a mathematical equation and
thus there is little space for ambiguity in the definition.

\paragraph{Objective.}
Universal intelligence does not depend on any subjective criteria.

\paragraph{Universal.}
Universal intelligence is in no way anthropocentric.

\paragraph{Practical.}
In its current form the definition cannot be directly turned into a
test of intelligence as the Kolmogorov complexity function is not
computable.  Thus in its pure form we can only use it to analyse the
nature of intelligence and to theoretically examine the intelligence
of mathematically defined learning algorithms.

In order to use universal intelligence more generally we will need to
construct a workable test that approximates an agent's $\Upsilon$
value.  The equation for $\Upsilon$ suggests how we might approach
this problem.  Essentially, an agent's universal intelligence is a
weighted sum of its performance over the space of all environments.
Thus, we could randomly generate programs that describe environmental
probability measures and then test the agent's performance against
each of these environments.  After sampling sufficiently many
environments the agent's approximate universal intelligence would be
computed by weighting its score in each environment according to the
complexity of the environment as given by the length of its program.
Another possibility might to be try to approximate the sum by
enumerating environmental programs from short to long, as the short
ones will contribute by far the greatest to the sum.  However in this
case we will need to be able to reset the state of the agent so that
it cannot cheat by learning our environmental enumeration method.  In
any case, various practical challenges will need to be addressed
before universal intelligence can be used to construct an effective
intelligence test.  As this would be a significant project in its own
right, in this paper we focus on the theoretical issues surrounding
the universal intelligence.

\section{Definitions and Tests of Machine Intelligence}\label{sec:ai}

In this section we will survey both definitions and tests of machine
intelligence.  We begin with a sample of informal definitions of
machine intelligence.  A comprehensive survey of informal definitions
is practically impossible as they often appear buried deep in articles
or books.  Nevertheless we have attempted to collect as many as
possible and present a sample of some of the more common perspectives
that have been taken.

We then survey formal definitions and tests of machine intelligence.
As we will see, it is not always clear whether a proposal is a test, a
formal definition, or something in between.  In some cases the authors
claim it is one or the other, and in some cases both.  More accurately
there is a spectrum of possibilities and thus we will not attempt to
artificially divide them into either tests or formal definitions.

To the best of our knowledge, this section is the only general survey
of tests and definitions of machine intelligence.  This is remarkable
given that the definition and measurement of intelligence in machines
are two of the most fundamental questions in artificial intelligence.
Currently most text books say very little about intelligence, other
than mentioning the Turing test.  We hope that our short survey will
help to raise awareness of the many other proposals.

\subsection{Informal definitions of machine intelligence}
\label{subsec:idmi}

The following sample of informal definitions of machine intelligence
capture a range of perspectives.  For a more comprehensive list of
definitions, visit our full collection online~\cite{Legg:07idefs}.  We
begin with five definitions that have clear connections to our
informal definition:

\begin{itemize}
\item[] ``\ldots the mental ability to sustain successful life.''
  K. Warwick quoted in~\cite{Asohan:03}
\end{itemize}

\begin{itemize}
\item[] ``\ldots doing well at a broad range of tasks is an empirical
  definition of `intelligence'$\;$'' H. Masum~\cite{Masum:02}
\end{itemize}

\begin{itemize}
\item[] ``Intelligence is the computational part of the ability to
  achieve goals in the world. Varying kinds and degrees of
  intelligence occur in people, many animals and some machines.''
  J. McCarthy~\cite{McCarthy:04}
\end{itemize}

\begin{itemize}
\item[] ``Any system \ldots that generates adaptive behaviour to meet
  goals in a range of environments can be said to be intelligent.''
  D. Fogel~\cite{Fogel:95}
\end{itemize}

\begin{itemize}
\item[] ``\ldots the ability of a system to act appropriately in an
  uncertain environment, where appropriate action is that which
  increases the probability of success, and success is the achievement
  of behavioral subgoals that support the system's ultimate goal.''
  J. S. Albus~\cite{Albus:91}
\end{itemize}

The position taken by Albus is especially similar to ours.  Although
the quote above does not explicitly mention the need to be able to
perform well in a wide range of environments, at a later point in the
same paper he mentions the need to be able to succeed in a ``large
variety of circumstances''.

\begin{itemize}
\item[] ``Intelligent systems are expected to work, and work well, in
  many different environments. Their property of intelligence allows
  them to maximize the probability of success even if full knowledge
  of the situation is not available.  Functioning of intelligent
  systems cannot be considered separately from the environment and the
  concrete situation including the goal.''
  R. R. Gudwin~\cite{Gudwin:00}
\end{itemize}

While this definition is consistent with the position we have taken,
when trying to actually test the intelligence of an agent Gudwin does
not believe that a ``black box'' behaviour based approach is
sufficient, rather his approach is to look at the ``\ldots
architectural details of structures, organizations, processes and
algorithms used in the construction of the intelligent systems,''
\cite{Gudwin:00} Our perspective is simply to not care whether an
agent looks intelligent on the inside.  If it is able to perform well
in a wide range of environments, that is all that matters.  For more
discussion on this point see our response to Block's and Searle's
arguments in Subsection~\ref{subsec:crit}.

\begin{itemize}
\item[] ``We define two perspectives on artificial system
  intelligence: (1) native intelligence, expressed in the specified
  complexity inherent in the information content of the system, and
  (2) performance intelligence, expressed in the successful (i.e.,
  goal-achieving) performance of the system in a complicated
  environment.''  J. A. Horst~\cite{Horst:02}
\end{itemize}

Here we see two distinct notions of intelligence, a performance based
one and an information content one.  This is similar to the
distinction between fluid intelligence and crystallized intelligence
made by the psychologist Cattell (see Subsection~\ref{subsec:toi}).
The performance notion of intelligence is similar to our definition
with the expectation that performance is measured in a complex
environment rather than across a wide range of environments.  This
perspective appears in some other definitions also,

\begin{itemize}
\item[] ``\ldots the ability to solve hard problems.''
  M. Minsky~\cite{Minsky:85}
\end{itemize}

\begin{itemize}
\item[] ``Achieving complex goals in complex environments''
  B. Goertzel~\cite{Goertzel:06}
\end{itemize}

Interestingly, Goertzel claims that an AI system he is developing
should be able to, with sufficient resources, perform arbitrarily well
with respect to the intelligence order relation, that is, the relation
on which universal intelligence was originally based~\cite{Looks:04}.
Presumably then he does not consider his definition to be
significantly incompatible with ours.

Some definitions emphasise not just the ability to perform well, but
also the need for efficiency:

\begin{itemize}
\item[] ``[An intelligent agent does what] is appropriate for its
  circumstances and its goal, it is flexible to changing environments
  and changing goals, it learns from experience, and it makes
  appropriate choices given perceptual limitations and finite
  computation.'' D. Poole~\cite{Poole:98}
\end{itemize}

\begin{itemize}
\item[] ``\ldots in any real situation behavior appropriate to the
  ends of the system and adaptive to the demands of the environment
  can occur, within some limits of speed and complexity.'' A. Newell
  and H. A. Simon~\cite{Newell:76}
\end{itemize}

\begin{itemize}
\item[] ``Intelligence is the ability to use optimally limited
  resources -- including time -- to achieve goals.''
  R. Kurzweil~\cite{Kurzweil:00}
\end{itemize}

\begin{itemize}
\item[] ``Intelligence is the ability for an information processing
  agent to adapt to its environment with insufficient knowledge and
  resources.'' P. Wang~\cite{Wang:95}
\end{itemize}

We consider the addition of resource limitations to the definition of
intelligence to be either superfluous, or wrong.  In the first case,
if limited computational resources are a fundamental and unavoidable
part of reality, which certainly seems to be the case, then their
addition to the definition of intelligence is unnecessary.  Perhaps
the first three definitions above fall into this category.

On the other hand, if limited resources are not a fundamental
restriction, for example a new model of computation was discovered
that was vastly more powerful than the current model, then it would be
odd to claim that the unbelievably powerful machines that would then
result were not intelligence.  Normally we do not judge the
intelligence of something relative to the resources it uses.  For
example, if a rat had human level learning and problem solving
abilities, we would not think of the rat as being more intelligence
than a human due to the fact that its brain was much smaller.

While we do not consider efficiency to be a part of the definition of
intelligence, this is not to say that considering the efficiency of
agents is unimportant.  Indeed, a key goal of artificial intelligence
is to find algorithms which have the greatest efficiency of
intelligence, that is, which achieve the most intelligence per unit of
computational resources consumed.

It should also be pointed out that although universal intelligence
does not test the efficiency of an agent in terms of the computational
resources that it uses, it does however test how quickly the agent
learns from past data.  In a sense, an agent which learns very quickly
could be thought of as being very ``data efficient''.

\subsection{Formal definitions and tests of machine intelligence}
\label{subsec:tmi}

\paragraph{Turing test and derivatives.}
The classic approach to determining whether a machine is intelligent
is the so called Turing test \cite{Turing:50} which has been
extensively debated over the last 50 years \cite{Saygin:00}.  Turing
realised how difficult it would be to directly definite intelligence
and thus attempted to side step the issue by setting up his now famous
imitation game: If human judges can not effectively discriminate
between a computer and a human through teletyped conversation then we
must conclude that the computer is intelligent.

Though simple and clever, the test has attracted much criticism.
Block and Searle argue that passing the test is not \emph{sufficient}
to establish intelligence \cite{Block:81,Searle:80,Eisner:91}.
Essentially they both argue that a machine could appear to be
intelligent without having any ``real intelligence'', perhaps by using
a very large table of answers to questions.  While such a machine
might be impossible in practice due to the vast size of the table
required, it is not logically impossible.  In which case an
unintelligent machine could, at least in theory, consistently pass the
Turing test.  Some consider this to bring the validity of the test
into question.  In response to these challenges, even more demanding
versions of the Turing test have been proposed such as the Total
Turing test \cite{Harnad:89}, the Truly Total Turing test
\cite{Schweizer:98} and the inverted Turing test \cite{Watt:96}.  Dowe
argues that the Turing test should be extended by ensuring that the
agent has a compressed representation of the domain area, thus ruling
out look-up table counter arguments~\cite{Dowe:98}.  Of course these
attacks on the Turing test can be applied to any test of intelligence
that considers only a system's external behaviour, that is, most
intelligence tests.

A more common criticism is that passing the Turing test is not
\emph{necessary} to establish intelligence.  Usually this argument is
based on the fact that the test requires the machine to have a highly
detailed model of human knowledge and patterns of thought, making it a
test of humanness rather than intelligence \cite{French:90,Ford:98}.
Indeed, even small things like pretending to be \emph{unable} to
perform complex arithmetic quickly and faking human typing errors
become important, something which clearly goes against the purpose of
the test.

The Turing test has other problems as well.  Current AI systems are a
long way from being able to pass an unrestricted Turing test.  From a
practical point of view this means that the full Turing test is unable
to offer much guidance to our work.  Indeed, even though the Turing
test is the most famous test of machine intelligence, almost no
current research in artificial intelligence is specifically directed
toward being able to pass it.  Unfortunately, simply restricting the
domain of conversation in the Turing test to make the test easier, as
is done in the Loebner competition \cite{Loebner:90}, is not
sufficient.  With restricted conversation possibilities the most
successful Loebner entrants are even more focused on faking human
fallibility, rather than anything resembling intelligence
\cite{Hutchens:96}. Finally, the Turing test returns different results
depending on who the human judges are.  Its unreliability has in some
cases lead to clearly unintelligent machines being classified as
human, and at least one instance of a human actually failing a Turing
test.  When queried about the latter, one of the judges explained that
``no human being would have that amount of knowledge about
Shakespeare''\cite{Shieber:94}.

\paragraph{Compression tests.}
Mahoney has proposed a particularly simple solution to the binary pass
or fail problem with the Turing test: Replace the Turing test with a
text compression test \cite{Mahoney:99}.  In essence this is somewhat
similar to a ``Cloze test'' where an individual's comprehension and
knowledge in a domain is estimated by having them guess missing words
from a passage of text.

While simple text compression can be performed with symbol
frequencies, the resulting compression is relatively poor.  By using
more complex models that capture higher level features such as aspects
of grammar, the best compressors are able to compress text to about
1.5 bits per character for English.  However humans, which can also
make use of general world knowledge, the logical structure of the
argument etc., are able to reduce this down to about 1 bit per
character.  Thus the compression statistic provides an easily computed
measure of how complete a machine's models of language, reasoning and
domain knowledge are, relative to a human.

To see the connection to the Turing test, consider a compression test
based on a very large corpus of dialogue.  If a compressor could
perform extremely well on such a test, this is mathematically
equivalent to being able to determine which sentences are probable at
a give point in a dialogue, and which are not (for the equivalence of
compression and prediction see~\cite{Bell:90}).  Thus, as failing a
Turing test occurs when a machine (or person!) generates a sentence
which would be improbable for a human, extremely good performance on
dialogue compression implies the ability to pass a Turing test.

A recent development in this area is the Hutter
Prize~\cite{Hutter:06hprize}.  In this test the corpus is a 100 MB
extract from Wikipedia.  The idea is that this should represent a
reasonable sample of world knowledge and thus any compressor that can
perform very well on this test must have have a good model of not just
English, but also world knowledge in general.

One criticism of compression tests is that it is not clear whether a
powerful compressor would easily translate into a general purpose
artificial intelligence.  Also, while a young child has a significant
amount of elementary knowledge about how to interact with the world,
this knowledge would be of little use when trying to compress an
encyclopedia full of abstract ``adult knowledge'' about the world.

\paragraph{Linguistic complexity.}
A more linguistic approach is taken by the HAL project at the company
Artificial Intelligence NV \cite{Treister:01}.  They propose to
measure a system's level of conversational ability by using techniques
developed to measure the linguistic ability of children.  These
methods examine things such as vocabulary size, length of utterances,
response types, syntactic complexity and so on.  This would allow
systems to be ``\ldots assigned an age or a maturity level beside
their binary Turing test assessment of `intelligent' or `not
intelligent'~''\cite{Treister:00}.  As they consider communication to
be the basis of intelligence, and the Turing test to be a valid test
of machine intelligence, in their view the best way to develop
intelligence is to retrace the way in which human linguistic
development occurs.  Although they do not explicitly refer to their
linguistic measure as a test of intelligence, because it measures
progress towards what they consider to be a valid intelligence test,
it acts as one.

\paragraph{Multiple cognitive abilities.}
A broader developmental approach is being taken by IBM's Joshua Blue
project \cite{Alvarado:02}.  In this project they measure the
performance of their system by considering a broad range of
linguistic, social, association and learning tests.  Their goal is to
first pass what they call a ``toddler Turing test'', that is, to
develop an AI system that can pass as a young child in a similar set
up to the Turing test.

Another company pursuing a similar developmental approach based on
measuring system performance through a broad range of cognitive tests
is the a2i2 project at Adaptive AI \cite{Voss:05}.  Rather than
toddler level intelligence, their current goal to is work toward a
level of cognitive performance similar to that of a small mammal.  The
idea being that even a small mammal has many of the key cognitive
abilities required for human level intelligence working together in an
integrated way.

\paragraph{Competitive games.}
The Turing Ratio method of Masum et al.\ has more emphasis on tasks
and games rather than cognitive tests.  Similar to our own definition,
they propose that ``\ldots doing well at a broad range of tasks is an
empirical definition of `intelligence'."\cite{Masum:02} To quantify
this they seek to identify tasks that measure important abilities,
admit a series of strategies that are qualitatively different, and are
reproducible and relevant over an extended period of time.  They
suggest a system of measuring performance through pairwise comparisons
between AI systems that is similar to that used to rate players in the
international chess rating system.  The key difficulty however, which
the authors acknowledge is an open challenge, is to work out what
these tasks should be, and to quantify just how broad, important and
relevant each is.  In our view these are some of the most central
problems that must be solved when attempting to construct an
intelligence test.  Thus we consider this approach to be incomplete in
its current state.

\paragraph{Collection of psychometric tests.}
An approach called Psychometric AI tries to address the problem of
what to test for in a pragmatic way.  In the view of Bringsjord and
Schimanski, ``Some agent is intelligent if and only if it excels at
all established, validated tests of [human]
intelligence.''\cite{Bringsjord:03} They later broaden this to also
include ``tests of artistic and literary creativity, mechanical
ability, and so on.''  With this as their goal, their research is
focused on building robots that can perform well on standard
psychometric tests designed for humans, such as the Wechsler Adult
Intelligent Scale and Raven Progressive Matrices (see
Subsection~\ref{subsec:sti}).

As effective as these tests are for humans, we believe that they are
unlikely to be adequate for measuring machine intelligence.  For a
start they are highly anthropocentric.  Another problem is that they
embody basic assumptions about the test subject that are likely to be
violated by computers.  For example, consider the fundamental
assumption that the test subject is not simply a collection of
specialised algorithms designed only for answering common IQ test
questions.  While this is obviously true of a human, or even an ape,
it may not be true of a computer.  The computer could be nothing more
than a collection of specific algorithms designed to identify patterns
in shapes, predict number sequences, write poems on a given subject or
solve verbal analogy problems --- all things that AI researchers have
worked on.  Such a machine might be able to obtain a respectable IQ
score \cite{Sanghi:03}, even though outside of these specific test
problems it would be next to useless.  If we try to correct for these
limitations by expanding beyond standard tests, as Bringsjord and
Schimanski seem to suggest, this once again opens up the difficulty of
exactly what, and what not, to test for.  Thus we consider
Psychometric AI, at least as it is currently formulated, to only
partially address this central question.

\paragraph{C-Test.}
One perspective among psychologists who support the $g$-factor view of
intelligence, is that intelligence is ``the ability to deal with
complexity''\cite{Gottfredson:97}.  Thus, in a test of intelligence,
the most difficult questions are the ones that are the most complex
because these will, by definition, require the most intelligence to
solve.  It follows then that if we could formally define and measure
the complexity of test problems using complexity theory we could
construct a formal test of intelligence.  The possibility of doing
this was perhaps first suggested by Chaitin \cite{Chaitin:82}.  While
this path requires numerous difficulties to be dealt with, we believe
that it is the most natural and offers many advantages: It is formally
motivated, precisely defined and potentially could be used to measure
the performance of both computers and biological systems on the same
scale without the problem of bias towards any particular species or
culture.

Essentially this is the approach that we have taken.  Universal
intelligence is based our the universally optimal AIXI agent for
active environments, which in turn is based on Kolmogorov complexity
and Solomonoff's universal model of sequence prediction.  A relative
of universal intelligence is the C-Test of Hern\'{a}ndez-Orallo which
was also inspired by Solomonoff induction and Kolmogorov
complexity~\cite{Hernandez:00cmi,Hernandez:98fdi}.  If we gloss over
some technicalities, the essential relationships look like this:\\

\begin{center}
\begin{tabular}{|c||c|c|} \hline
                           & \emph{Universal agent} & \emph{Universal test} \\
\hline\hline
\emph{Passive environment} & Solomonoff induction & C-Test \\
\hline
\emph{Active environment}  & AIXI                 & Universal intelligence \\
\hline
\end{tabular}
\end{center}

\vspace{1.5em}

The C-Test consists of a number of sequence prediction and abduction
problems similar to those that appear in many standard IQ tests.  The
test has been successfully applied to humans with intuitively
reasonable results \cite{Hernandez:98fdi,Hernandez:00btt}.  Similar to
standard IQ tests, the C-Test always ensures that each question has an
unambiguous answer in the sense that there is always one hypothesis
that is consistent with the observed pattern that has significantly
lower complexity than the alternatives.  Other than making the test
easier to score, it has the added advantage of reducing the test's
sensitivity to changes in the reference machine.

The key difference to sequence problems that appear in standard
intelligence tests is that the questions are based on a formally
expressed measure of complexity.  To overcome the problem of
Kolmogorov complexity not being computable, the C-Test instead uses
Levin's $Kt$ complexity \cite{Levin:73search}.  In order to retain the
invariance property of Kolmogorov complexity, Levin complexity
requires the additional assumption that the universal Turing machines
are able to simulate each other in linear time.  As far as we know,
this is the only formal definition of intelligence that has so far
produced a usable test of intelligence.

To illustrate the C-Test, below are some example problems taken
from~\cite{Hernandez:98fdi}.  Beside each question is its complexity,
naturally more complex patterns are also more difficult:\\

\begin{center}
\begin{tabular}{cp{6cm}c}
& \emph{Sequence Prediction Test} \\[0.5ex]
Complexity & Sequence & Answer\\
 9 & a, d, g, j, \_ , \ldots & m \\
12 & a, a, z, c, y, e, x, \_ , \ldots & g \\
14 & c, a, b, d, b, c, c, e, c, d, \_ , \ldots & d \\
\end{tabular}

\vspace{1.5em}

\begin{tabular}{cp{6cm}c}
& \emph{Sequence Abduction Test} \\[0.5ex]
Complexity & Sequence & Answer\\
 8 & a, \_ , a, z, a, y, a, \ldots & a \\
10 & a, x, \_ , v, w, t, u, \ldots & y \\
13 & a, y, w, \_ , w, u, w, u, s, \ldots & y \\
\end{tabular}
\end{center}

\vspace{1.5em}

Our main criticism of the C-Test is that it is a static test limited
to passive environments.  As we have argued earlier, we believe that a
better approach is to use dynamic intelligence tests where the agent
must interact with an environment in order to solve problems.  As AIXI
is a generalisation of Solomonoff induction from passive to active
environments, universal intelligence could be viewed as generalising
the C-Test from passive to active environments.

\paragraph{Smith's Test.}
Another complexity based formal definition of
intelligence that appeared recently in an unpublished report is due to
W. D. Smith~\cite{Smith:06}.  His approach has a number of connections
to our work, indeed Smith states that his work is largely a ``\ldots
rediscovery of recent work by Marcus Hutter''.  Perhaps this is over
stating the similarities because while there are some connections,
there are also many important differences.

The basic structure of Smith's definition is that an agent faces a
series of problems that are generated by an algorithm.  In each
iteration the agent must try to produce the correct response to the
problem that it has been given.  The problem generator then responds
with a score of how good the agent's answer was.  If the agent so
desires it can submit another answer to the same problem.  At some
point the agent requests to the problem generator to move onto the
next problem and the score that the agent received for its last answer
to the current problem is then added to its cumulative score.  Each
interaction cycle counts as one time step and the agent's intelligence
is then its total cumulative score considered as a function of time.
In order to keep things feasible, the problems must all be in the
complexity class P, that is, decision problems which can be solved by
a deterministic Turing machine in polynomial time.

We have three main criticisms of Smith's definition.  Firstly, while
for practical reasons it might make sense to restrict problems to be
in P, we do not see why this practical restriction should be a part
of the very definition of intelligence.  If some breakthrough meant
that agents could solve difficult problems in not just P but
sometimes in NP as well, then surely these new agents would be more
intelligent?  We had similar objections to informal definitions of
machine intelligence that included efficiency requirements in
Subsection~\ref{subsec:idmi}.

Our second criticism is that the way intelligence is measured is
essentially static, that is, the environments are passive.  As we have
argued before, we believe that dynamic testing in active environments
is a better measure of a system's intelligence.  To put this argument
yet another way: Succeeding in the real world requires you to be more
than an insightful spectator!

The final criticism is that while the definition is somewhat formally
defined, still it leaves open the important question of what exactly
the tests should be.  Smith suggests that researchers should dream up
tests and then contribute them to some common pool of tests.  As such,
this is not a fully specified definition.

\subsection{Comparison of machine intelligence tests and definitions}
\label{subsec:despropmi}

In order to compare the machine intelligence tests and definitions in
the previous section, we return again to the desirable properties of a
test of intelligence.

Each property is briefly defined followed by a summary comparison in
Table~\ref{table:mitests}.  Although we have attempted to be as fair
as possible, some of the scores we give on this table will be
debatable.  Nevertheless, we hope that it provides a rough overview of
the relative strengths and weaknesses of the proposals.

\begin{description}

\item{\bf Valid.} A test/measure of intelligence should be just that,
  it should capture intelligence and not some related quantity or only
  a part of intelligence.

\item{\bf Informative.} The result should be a scalar value, or
  perhaps a vector, depending on our view of intelligence.  We would
  like an absolute measure of intelligence so that comparisons across
  many agents can easily be made.

\item{\bf Wide range.} A test/definition should cover very low levels
  of intelligence right up to super human intelligence.

\item{\bf General.} Ideally we would like to have a very general
  test/definition that could be applied to everything from a fly to a
  machine learning algorithm.

\item{\bf Dynamic.} A test/definition should directly take into
  account the ability to learn and adapt over time as this is an
  important aspect of intelligence.

\item{\bf Unbiased.} A test/definition should not be biased towards
  any particular culture, species, etc.

\item{\bf Fundamental.} We do not want a test/definition that needs to
  be changed from time to time due to changing technology and
  knowledge.

\item{\bf Formal.} The test/definition should be specified with the
  highest degree of precision possible, allowing no room for
  misinterpretation.  Ideally, it should be described using formal
  mathematics.

\item{\bf Objective.} The test/definition should not appeal to
  subjective assessments such as the opinions of human judges.

\item{\bf Fully Defined.} Has the test/definition been fully defined,
  or are parts still unspecified?

\item{\bf Universal.} Is the test/definition universal, or is it
  anthropocentric?

\item{\bf Practical.} A test should be able to be performed quickly
  and automatically, while from a definition it should be possible to
  create an efficient test.

\item{\bf Test vs.\ Def.}  Finally, we note whether the proposal is more
  of a test, more of a definition, or something in between.

\end{description}


\begin{center}
\begin{table}
\begin{tabular}{l|c|c|c|c|c|c|c|c|c|c|c|c|c|}
 \\
\multicolumn{1}{l|}{Intelligence Test} \\
&
\multicolumn{1}{c}{\begin{rotate}{45}Valid\end{rotate}} &
\multicolumn{1}{c}{\begin{rotate}{45}Informative\end{rotate}} &
\multicolumn{1}{c}{\begin{rotate}{45}Wide Range\end{rotate}} &
\multicolumn{1}{c}{\begin{rotate}{45}General\end{rotate}} &
\multicolumn{1}{c}{\begin{rotate}{45}Dynamic\end{rotate}} &
\multicolumn{1}{c}{\begin{rotate}{45}Unbiased\end{rotate}} &
\multicolumn{1}{c}{\begin{rotate}{45}Fundamental\end{rotate}} &
\multicolumn{1}{c}{\begin{rotate}{45}Formal\end{rotate}} &
\multicolumn{1}{c}{\begin{rotate}{45}Objective\end{rotate}} &
\multicolumn{1}{c}{\begin{rotate}{45}Fully Defined\end{rotate}} &
\multicolumn{1}{c}{\begin{rotate}{45}Universal\end{rotate}} &
\multicolumn{1}{c}{\begin{rotate}{45}Practical\end{rotate}} &
\multicolumn{1}{c}{\begin{rotate}{45}Test vs.\ Def.\end{rotate}} \\
\hline
Turing Test            &\tB & \tC &\tC &\tC &\tB &\tC &\tC &\tC &\tC &\tB &\tC &\tB & T \\
Total Turing Test      &\tB & \tC &\tC &\tC &\tB &\tC &\tC &\tC &\tC &\tB &\tC &\tC & T \\
Inverted Turing Test   &\tB & \tB &\tC &\tC &\tB &\tC &\tC &\tC &\tC &\tB &\tC &\tB & T \\
Toddler Turing Test    &\tB & \tC &\tC &\tC &\tB &\tC &\tC &\tC &\tC &\tC &\tC &\tB & T \\
Linguistic Complexity  &\tB & \tA &\tB &\tC &\tC &\tC &\tC &\tB &\tB &\tC &\tB &\tB & T \\
Text Compression Test  &\tB & \tA &\tA &\tB &\tC &\tB &\tB &\tA &\tA &\tA &\tB &\tA & T \\
Turing Ratio           &\tB & \tA &\tA &\tA & ?  & ?  & ?  & ?  & ?  &\tC & ?  & ?  & T/D \\
Psychometric AI        &\tA & \tA &\tB &\tA & ?  &\tB &\tC &\tB &\tB &\tB &\tC &\tB & T/D \\
Smith's Test           &\tB & \tA &\tA &\tB &\tC & ?  &\tA &\tA &\tA &\tC & ?  &\tB & T/D \\
C-Test                 &\tB & \tA &\tA &\tB &\tC &\tA &\tA &\tA &\tA &\tA &\tA &\tA & T/D \\
Universal Intelligence &\tA & \tA &\tA &\tA &\tA &\tA &\tA &\tA &\tA &\tA &\tA &\tC & D \\
\cline{2-14}
\end{tabular}
\caption{\label{table:mitests} In the table \tA\ means ``yes'',
  \tB\ means ``debatable'', \tC\ means ``no'', and ? means unknown.
  When something is rated as unknown that is usually because the test
  in question is not sufficiently specified.}
\end{table}
\end{center}

\section{Discussion and Conclusions}\label{sec:conc}

\subsection{Constructing a test of universal intelligence}

The central challenge for future work on universal intelligence is to
convert the theoretical definition of machine intelligence presented
in this paper into a workable test.  The basic structure of such a
test is already apparent from the equation for $\Upsilon$: The test
would work by evaluating the performance of an agent on a large sample
of simulated environments, and then combining the agent's performance
in each environment into an overall intelligence value.  This would be
done by weighting the agent's performance in each environment
according to the environment's complexity.

The key theoretical challenge that will need to be deal with is to
find a suitable replacement for the incomputable Kolmogorov complexity
function.  One solution could be to use Levin's $Kt$ complexity
\cite{Levin:73search}, another might be to use Schmidhuber's Speed
prior \cite{Schmidhuber:02speed}.  Both of these consider the
complexity of an algorithm to be determined by both its minimal
description length and running time.  This forces the complexity
measures to be computable.  Taking computation time into account also
makes reasonable intuitive sense because we would not usually consider
a very short algorithm that takes an enormous amount of time to run to
be a particularly simple one.  The fact that such an approach can be
made to work is evidenced by the C-Test.

\subsection{Response to common criticisms}
\label{subsec:crit}

What we have attempted to do is very ambitious and so, not
surprisingly, the reactions we get can be interesting.  Having
presented the essence of this work as posters at several conferences,
and also as a 30 minute talk, we now have some idea of what the
typical responses are.  Most people start out skeptical but end up
generally enthusiastic, even if they still have a few reservations.
This positive feedback has helped motivate us to continue this
direction of research.  In this subsection, however, we will attempted
to cover some of the more common criticisms.

\paragraph{It's obviously false, there's nothing in your definition, just a few equations.}
Perhaps the most common criticism is also the most vacuous one: It's
obviously wrong!  These people seem to believe that defining
intelligence with an equation is clearly impossible, and thus there
must be very large and obvious flaws in our work.  Not surprisingly
these people are also the least likely to want to spend 10 minutes
having the material explained to them.  Unfortunately, none of these
people have been able to communicate why the work is so obviously
flawed in any concrete way --- despite in one instance having one of
the authors chasing the poor fellow out of the conference centre and
down the street begging for an explanation.  If anyone would like to
properly explain their position to us in the future, we promise not to
chase you down the street!

\paragraph{It's obviously correct, indeed everybody already knows this stuff.}
Curiously, the second most common criticism is the exact opposite: The
work is obviously right, and indeed it is already well known.  Digging
deeper, the heart of this criticism comes from the perception that we
have not done much more than just describe reinforcement learning.  If
you already accept that the reinforcement learning framework is the
most general and flexible way to describe artificial intelligence, and
not everybody does, then by mixing in Occam's razor and a dash of
complexity theory, the equation for universal intelligence follows in
a fairly straightforward way.  While this is true, the way in which we
have brought these things together has never been done before,
although it does have some connection to other work, as discussed in
Subsection~\ref{subsec:tmi}.  Furthermore, simply coming up with an
equation is not enough, one must argue that what the equation
describes is in fact ``intelligence'' in a sense that is reasonable
for machines.

We have addressed this question in three main ways: Firstly, in
Section~\ref{sec:ni} we developed an informal definition of
intelligence based on expert definitions which was then piece by piece
formalised leading to the equation for $\Upsilon$ in
Subsection~\ref{subsec:fmi}.  This chain of argument strongly ties our
equation for intelligence with existing informal definitions and ideas
on the nature of intelligence.  Secondly, in
Subsections~\ref{subsec:eaae} and \ref{subsec:anal} we showed that the
equation has properties that are consistent with a definition of
intelligence.  Finally, in Subsection~\ref{subsec:eaae} it was shown
that universal intelligence is strongly connected to the theory of
universally optimal learning agents, in particular AIXI.  From this it
follows that machines with very high universal intelligence have a
wide range of powerful optimality properties.  Clearly then, what we
have done goes far beyond merely restating elementary reinforcement
learning theory.

\paragraph{Assuming that the environment is computable is too strong.}
It is certainly possible that the physical universe is not computable,
in the sense that the probability distribution over future events
cannot, even in theory, be simulated to an arbitrary precision by a
computable process.  Some people take this position on various
philosophical grounds, such as the need for freewill.  However, in
standard physics there is no law of the universe that is not
computable in the above sense.  Nor is there any experimental evidence
showing that such a physical law must exist.  This includes quantum
theory and chaotic systems, both of which can be extremely difficult
to compute for some physical systems, but are not fundamentally
incomputable theories.  In the case of quantum computers, they can
compute with lower time complexity than classical Turing machines,
however they are unable to compute anything that a classical Turing
machine cannot, when given enough time.  Thus, as there is no hard
evidence of incomputable processes in the universe, our assumption
that the agent's environment has a computable distribution is
certainly not unreasonable.

If a physical process was ever discovered that was not Turing
computable, then this would likely result in a new extended model of
computation.  Just as we have based universal intelligence on the
Turing model of computation, it might be possible to construct a new
definition of universal intelligence based on this new model in a
natural way.

Finally, even if the universe was not computable, and we did not
update our formal definition of intelligence to take this into
account, the fact that everything in physics so far is computable
means that a computable approximation to our universe would still be
extremely accurate over a huge range of situations.  In which case, an
agent that could deal with a wide range of computable environments
would most likely still function well within such a universe.

\paragraph{Assuming that environments return bounded sum rewards is unrealistic.}
If an environment $\mu$ is an artificial game, like chess, then it
seems fairly natural for $\mu$ to meet any requirements in its
definition, such as having a bounded reward sum.  However if we think
of the environment $\mu$ as being the universe in which the agent
lives, then it seems unreasonable to expect that it should be required
to respect such a bound.

Strictly speaking, reward is an interpretation of the state of the
environment.  In this case the environment is the universe, and
clearly the universe does not have any notion of reward for particular
agents.  In humans this interpretation is internal, for example, the
pain that is experienced when you touch something hot.  In which case,
maybe it should really be a part of the agent rather than the
environment?  If we gave the agent complete control over rewards then
our framework would become meaningless: the perfect agent could simply
give itself constant maximum reward.  Perhaps the analogous situation
for humans would be taking drugs.

A more accurate framework would consist of an agent, an environment
and a separate goal system that interpreted the state of the
environment and rewarded the agent appropriately.  In such a set up
the bounded rewards restriction would be a part of the goal system and
thus the above philosophical problem would not occur.  However, for
our current purposes, it is sufficient just to fold this goal
mechanism into the environment and add an easily implemented
constraint to how the environment may generate rewards.  One simple
way to bound an environment's total rewards would be to use geometric
or harmonic discounting.

\paragraph{How do you respond to Block's ``Blockhead'' argument?}
The approach we have taken is unabashedly functional.  Theoretically,
we desired to have a formal, simple and very general definition.  This
is easier to do if we abstract over the internal workings of the agent
and define intelligence only in terms of external communications.
Practically, what matters is \emph{how well something works}.  By
definition, if an agent has a high value of $\Upsilon$, then it must
work well over a wide range of environments.

Block attacks this perspective by describing a machine that appears to
be intelligent as it is able to pass the Turing test, but is in fact
no more than just a big look-up table of questions and answers
\cite{Block:81} (for a related argument see \cite{Gunderson:71}).
Although such a look-up table based machine would be unfeasibly large,
the fact that a finite machine could in theory consistently pass the
Turing test, seemingly without any real intelligence, is worrisome.
Our formal measure of machine intelligence could be challenged in the
same way, as could any test of intelligence that relies only on an
agent's external behaviour.

Our response to this is very simple: If an agent has a very high value
of $\Upsilon$ then it is, by definition, able to successfully operate
in a wide range of environments.  We simply do not care whether the
agent is efficient, due to some very clever algorithm, or absurdly
inefficient, for example by using an unfeasibly gigantic look-up table
of precomputed answers.  The important point for us is that the
machine has an amazing ability to solve a huge range of problems in a
wide variety of environments.

\paragraph{How do you respond to Searle's ``Chinese room'' argument?}
Searle's Chinese room argument attacks our functional position in a
similar way by arguing that a system may appear to be intelligent
without really understanding anything \cite{Searle:80}.  From our
perspective, whether or not an agent understands what it is doing is
only important to the extent that it affects the measurable
performance of the agent.  If the performance is identical, as Searle
seems to suggest, then whether or not the room with Searle inside
understands the meaning of what is going on is of no practical
concern; indeed it is not even clear to us how to define
``understanding'' if its presence has no measurable effects.  So long
as the system as a whole has the powerful properties required for
universal machine intelligence, then we have the kind of extremely
general and powerful machine that we desire.  On the other hand, if
``understanding'' does have a measurable impact on an agent's
performance in some well defined situations, then it is of interest to
us.  In which case, because $\Upsilon$ measures performance in all
well defined situations, it follows that $\Upsilon$ is in part a
measure of how much understanding an agent has.

\paragraph{But you don't deal with consciousness (or creativity,
imagination, freewill, emotion, love, soul, etc.)}
We apply the same argument to consciousness, emotions, freewill,
creativity, the soul and other such things.  Our goal is to build
powerful and flexible machines and thus these somewhat vague
properties are only relevant to our goal to the extent to which they
have some measurable effect on performance in some well defined
environment.  If no such measurable effect exists, then they are not
relevant to our objective.  Of course this is not the same as saying
that these things do not exist.  The question is whether they are
relevant or not.  We would consider creativity, appropriately
defined, to have a significant impact on an agent's ability to adapt
to challenging environments.  Perhaps the same is also true of
emotions, freewill and other qualities.

\paragraph{Universal intelligence is impossible due to the No-Free-Lunch theorem.}
Some, such as Edmonds~\cite{Edmonds:06}, argue that universal
intelligence is impossible due to Wolpert's so called ``No Free
Lunch'' theorem \cite{Wolpert:97}.  However this theorem, or any of
the standard variants on it, cannot be applied to universal
intelligence for the simple reason that we have not taken a uniform
distribution over the space of environments.

It is conceivable that there might exist some more general kind of
``No Free Lunch'' theorem for agents that limits their maximal
intelligence according to our definition.  Clearly any such result
would have to apply only to computable agents given that the
incomputable AIXI agent faces no such limit.  If such a result were
true, it would suggest that our definition of intelligence is perhaps
too broad in its scope.  Currently we know of no such result.

Interestingly, if it could be shown that an upper limit on $\Upsilon$
existed for feasible machines and that humans performed above this
limit, then this would prove that humans have some incomputable
element to their operation, perhaps consciousness, which is of real
practical significance to their performance.

\subsection{Conclusion}

\begin{quote}\it
``\ldots we need a definition of intelligence that is applicable to
  machines as well as humans or even dogs.  Further, it would be
  helpful to have a relative measure of intelligence, that would
  enable us to judge one program more or less intelligent than
  another, rather than identify some absolute criterion.  Then it will
  be possible to assess whether progress is being made \ldots''
  \par\hfill --- {\sl W. L. Johnson~\cite{Johnson:92}}
\end{quote}

Given the obvious significance of formal definitions of intelligence
for research, and calls for more direct measures of machine
intelligence to replace the problematic Turing test and other
imitation based tests, little work has been done in this area.  In
this paper we have attempted to tackle this problem by taking an
informal definition of intelligence modelled on expert definitions of
human intelligence, and then generalise and formalise it.  We believe
that the resulting mathematical definition captures the concept of
machine intelligence in a very powerful and yet elegant way.
Furthermore, by considering alternative, more tractable measures of
complexity, practical tests that estimate universal intelligence
should be possible.  Developing such tests will be the next major task
in this direction of research.

The fact that we have stated our definition of machine intelligence in
precise mathematical terms, rather than the more usual vaguely worded
descriptions, means that there is no reason why criticisms of our
approach should not be equally clear and precise.  At the very least
we hope that this in itself will help raise the debate over the
definition and nature of machine intelligence to a new level of
scientific rigour.

\subsection*{Acknowledgements}

This work was supported by the Swiss NSF grant 200020-107616.

\bibliographystyle{alpha}
\addcontentsline{toc}{section}{\refname}

\begin{small}
\newcommand{\etalchar}[1]{$^{#1}$}

\end{small}

\end{document}